\documentclass[twoside]{article}
\usepackage{palatino} 
\usepackage{coli} 
\usepackage{fullname} 

\usepackage{here,qtree,pst-tree,multicol,setspace}
\usepackage{latexsym,amsmath,amssymb,array,psfig}
\usepackage{algorithmicx}
\usepackage[noend]{algpseudocode}

\newcommand{\ignore}[1]{}
\newcommand{\bitem}{\begin{itemize}} 
\newcommand{\eitem}{\end{itemize}}
\newcommand{\be}{\begin{enumerate}}
\newcommand{\ee}{\end{enumerate}}

\newcommand{\la}{\langle}
\newcommand{\ra}{\rangle}
\newcommand{\ba}[1]{\begin{array}{#1}}
\newcommand{\ea}{\end{array}}

\newcommand{\der}{\Rightarrow}

\issue{x}{y}{z}

\title{
Statistical Machine Translation by Generalized Parsing\thanks{A
preliminary version of this article was published by
Melamed (2004b).  This article improves and extends every section of that
preliminary version, and adds several new sections.}  }

\author{ I. Dan Melamed \\%
         \affil{New York University}
\and
Wei Wang \\
\affil{Language Weaver Inc.}
}

\runningtitle{Translation by generalized parsing 
}
\runningauthor{Melamed \& Wang}

\begin{document}

\begin{spacing}{1}



\maketitle

\begin{abstract}
Designers of statistical machine translation (SMT) systems have begun
to employ tree-structured translation models.  Systems involving
tree-structured translation models tend to be complex.  This article
aims to reduce the conceptual complexity of such systems, in order to
make them easier to design, implement, debug, use, study, understand,
explain, modify, and improve.  In service of this goal, the article
extends the theory of semiring parsing to arrive at a novel abstract
parsing algorithm with five functional parameters: a logic, a grammar,
a semiring, a search strategy, and a termination condition.  The
article then shows that all the common algorithms that revolve around
tree-structured translation models, including hierarchical alignment,
inference for parameter estimation, translation, and structured
evaluation, can be derived by generalizing two of these parameters ---
the grammar and the logic.  The article culminates with a recipe for
using such generalized parsers to train, apply, and evaluate an SMT
system that is driven by tree-structured translation models.
\end{abstract}

\section{Introduction}
\label{sec::intro}

\begin{figure}
\mbox{\psfig{figure=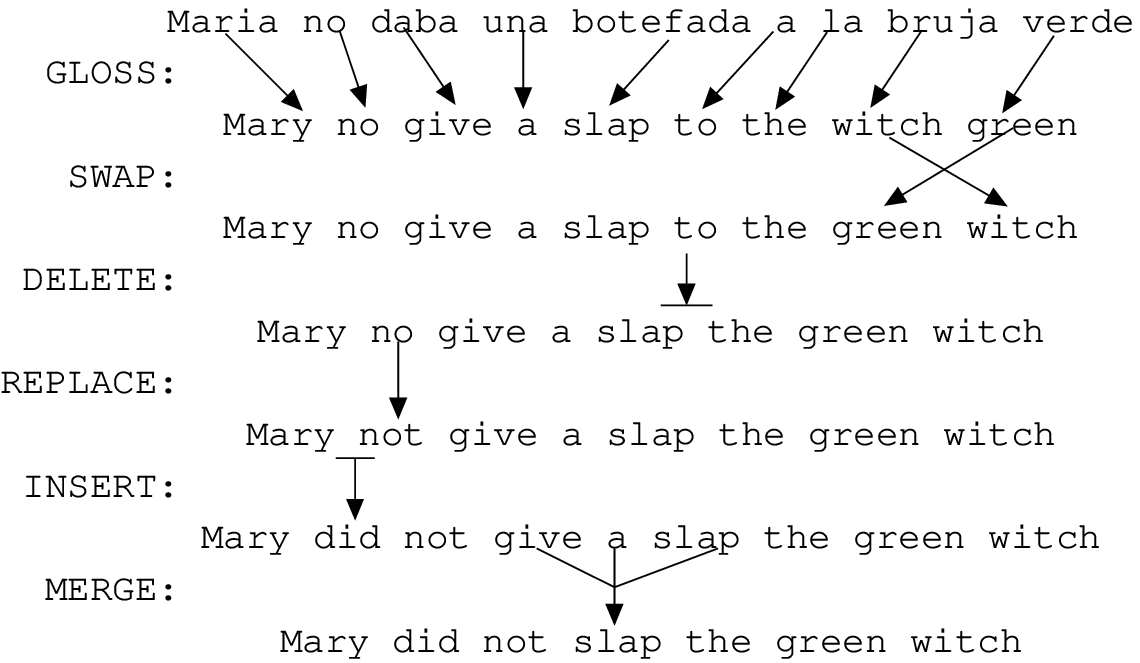,width=3in}}
\caption{}
{Translation by finite-state transduction.  Adapted from
\namecite{KnightKoehn03}.}
\label{fst}
\end{figure}

Today's best statistical machine translation (SMT) systems are driven
by translation models that are weighted finite-state transducers (WFSTs)
\cite{Och02,KumarByrne03}.  Figure~\ref{fst} shows a typical example
of a WFST translation model, and the way it is composed of a series of
sub-transducers.  Models of this type and our methods for using them
have become increasingly sophisticated in recent years, leading to
steady advances in the accuracy of the best MT systems.  However, such
translation models run counter to our intuitions about how expressions
in different languages are related.  In the short term, SMT research
based on WFSTs may be a necessary stepping stone, and it is still
possible to make improvements by hill-climbing on objective criteria.
In the long term, the price of implausible models is reduced insight.
There is a growing awareness in the SMT research community that major
advances can come only from deeper intuitions about the relationship of
our models to the phenomena being modeled.

From an engineering point of view, modeling translational equivalence
using WFSTs is like approximating a high-order polynomial with line
segments.  Given enough parameters, the approximation can be
arbitrarily good.  In practice, the number of parameters that can be
reliably estimated is limited either by the amount of available
training data or by the available computing resources.  Suitable
training data will always be limited for most of the world's
languages.  On the other hand, for resource-rich language pairs where
the amount of training data is practically infinite, the limiting
factor is the number of model parameters that fit into our computers'
memories.  Either way, the relatively low expressive power of WFSTs
limits the quality of SMT systems.

To advance the state of the art, SMT system designers have begun to
experiment with tree-structured translation models
\cite[e.g.]{Wu96,Alshawi96,YamadaKnight02,Gildea03,Chiang05}.  
Tree-structured translation models have the potential to encode more
information using fewer parameters.  For example, suppose we wish to
express the preference for adjectives to follow nouns in language L1
and to precede them in language L2.  A model that knows about parts of
speech needs only one parameter to record this binary preference.
Some finite-state translation models can encode parts of speech and
other word classes \cite{Och99}.  However, they cannot encode the
preferred relative order of noun {\em phrases} and adjectival {\em
phrases}, because this kind of knowledge involves parameters over
recursive structures.  To encode such knowledge, a model must be at
least tree-structured.  For example, a syntax-directed transduction
schema (SDTS) \cite{AhoUllman69} needs only one parameter to know
that an English noun phrase of the form (Det AdjP N) such as ``the
green and blue shirt'' translates to Arabic in the order (Det N AdjP).
A well-known principle of machine learning is that, everything else
being equal, models with fewer parameters are more likely to make
accurate predictions on previously unseen data.


Several authors have added tree-structured information to systems that
were primarily based on WFSTs \cite{Koehn+03,Eng+03}.  Such a system
can be easier to build, especially given pre-existing software for
WFST-based SMT.  However, such a system cannot reach the potential
parameter efficiency of a tree-structured translation model, because
it is still saddled with the large number of parameters required by
the underlying WFSTs.  Such hybrid systems are improving all the time.
Yet, one cannot help but wonder how much faster they would improve if
they were to shed their historical baggage. 

To realize the full potential of tree-structured models, an SMT system
must use them as the primary models in every stage of its operation,
including training, application to new inputs, and evaluation.
Switching to a less efficient model at any stage can result in an
explosion in the number of parameters necessary to encode the same
information.  If the resulting model no longer fits in memory, then
the system is forced to lose information, and thus also accuracy.
\footnote{An alternative is to swap the model out to secondary
storage, slowing down the system by several orders of magnitude.} Even
when memory is not an issue, the increased number of parameters risks
an increase in generalization error.

For these reasons, among others, the SMT research community is highly
motivated to build systems whose every process is driven primarily by
tree-structured models.  Unfortunately, from a naive point of view,
such systems tend to be conceptually complex --- much more complex
than WFST-based systems.
Complex systems present significant obstacles to research:
\bitem
\item They take a long time to design, to implement, to debug
and to document.  Given the fast pace of our field of research, good
software engineering is usually postponed until after the next
conference paper deadline.  So, typical implementations are difficult
to modify and to extend even for their authors, let alone for anybody
else.
\item The large number of possible variations in the algorithms
involved and in their parameters makes it difficult to run controlled
experiments. The large number of independent variables makes it
difficult to assign credit/blame for changes in the system's
accuracy.  Improvements are typically obtained by trial and error,
and followed by post-hoc explanations that may or may not be
scientifically valid.
\item The large number of variables that can affect the outcome of an
experiment make the experiments difficult to describe in detail.
Experiments that are not fully described are difficult to replicate.
Perhaps this is why most of the literature to date on tree-structured
translation models compares those models only to variations of
themselves and to WFST-based models, but not to other tree-structured
models in the literature.
\eitem
Despite the fast pace of research in this area, it is likely that
research would progress more quickly if it were not hindered by
the above obstacles.

The primary aim of this article is to reduce the conceptual
complexity of SMT systems driven by tree-structured translation
models, and thereby to reduce the obstacles outlined above.  In
service of this goal, Section~\ref{sec:anat} extends the theory of
semiring parsing to arrive at a novel analysis of many common parsing
algorithms.  This analysis led to two insights, which are expounded in
Sections~\ref{sec:mp} through~\ref{sec:eval}.  First, under a certain
parameterization, all of the non-trivial algorithms that are necessary
for this approach to SMT are special cases of just one algorithm.
Second, the one key algorithm that is necessary for this type of SMT
is a direct generalization of ordinary parsing. These insights imply
that:
\bitem
\item Implementation of an SMT system driven by tree-structured
translation models requires only one non-trivial software module.  The
software engineering effort of the implementation, as well as of any
subsequent extensions, is thereby reduced by an order of magnitude.
This reduction in effort makes the enterprise feasible for a much
larger number of researchers.  The ``Statistical Machine Translation
by Parsing'' Team at the 2005 JHU Language Engineering Workshop took
advantage of this new-found feasibility to build the first publicly
available toolkit for machine translation by parsing.\footnote{See
{\tt http://www.clsp.jhu.edu/ws2005/groups/statistical/GenPar.html} .}
\item An innovation or improvement in one algorithm will often be
applicable to all the others.  Conversely, a deeper understanding of the
relationships among these algorithms can lead to new insights about
the whole class.
\item Many of the problems that SMT research  will
encounter can be solved by generalizing existing solutions from the
parsing literature.  Such generalizations typically require less
effort than completely new solutions, as this article shall
demonstrate.
\eitem
The article makes no empirical claims about the merits of
tree-structured translation models or translation by parsing.
Instead, the aim is to reduce the effort necessary for research into
what those merits might be.

\section{Anatomy of a Parser}

\label{sec:anat}

In natural language processing, a parser is an algorithm for inferring
linguistic structure.  We limit our attention to parsers that infer
structure incrementally using a grammar \cite{JurafskyMartin00},
rather than by reranking a list of pre-existing structures
\cite{CollinsKoo05}, or by inferring an entire parse tree as a point
in a high-dimensional feature space
\cite{Taskar+04}.\footnote{At the time of writing, parsing by
structured classification is too expensive to train for practical
purposes, and reranking approaches rely on the kind of parsers that we
focus on.}  However, the grammar need not be generative or
probabilistic.  Our only requirement for the grammar is that it should
assign values to parts of parse tree structures.  These values can
range over booleans (structure is possible or not), probabilities,
feature weights \cite{Chiang05}, or other values such as confidence
estimates \cite{TurianMelamed05}.

To facilitate our generalization of ordinary parsers to algorithms
necessary for SMT, we shall recast them in terms of an abstract
parsing algorithm with five functional parameters: a grammar, a logic,
a semiring, a search strategy, and a termination condition.  We shall
then express all the algorithms necessary for SMT by generalizing two
of those parameters --- the grammar and the logic.  This
characterization will make it easier to compare and contrast these
algorithms.  The use of logics to describe parsers is not new (e.g.\
\namecite[and references therein]{Shieber+95}).
\namecite{KleinManning03} have compared different search strategies
for a fixed parsing logic and grammar. The parameterization of parsing
algorithms by semirings was studied by \namecite{Goodman98}, who also
presented an abstract parsing algorithm.  The abstract parsing
algorithm in Section~\ref{sec:apa} is more detailed and more
general.  Before presenting this algorithm, we shall explain some of
its parameters.  We presume that readers are already familiar with
probabilistic context-free grammars and ordinary parsers
\cite{JurafskyMartin00}.

\subsection{Logics for Parsing}

\label{sec:inf}

\label{sec:naive}

A parser's {\bf logic} determines the parser's possible states and
transitions.  The specification of a parsing logic has three parts:
\bitem
\item a set of term type signatures,
\item a set of inference rule type signatures,
\item a set of axiom terms.
\eitem
Terms are the building blocks of inference rules.  {\bf Items} are
terms that represent partial parses.  Terms that represent grammatical
constraints such as production rules are sometimes called {\bf grammar
terms}.
When the parser runs, the term and inference rule types are
instantiated and their variables are assigned values.  
The state of a parser can be uniquely specified by the values of all
possible terms.  

In the parser's initial state, all terms have a particular default
value, such as ``false'' or ``zero probability,'' depending on the
semiring (see below).  {\bf Axioms} are term instances (not types)
that are assigned non-default values during the parser's
initialization procedure.  The most common kinds of axioms come from
the grammar and from the input.  Typically, each input word gives
rise to an axiom.  If the grammar involves production rules, then each
production rule becomes an axiom too. As a parser runs, it can change
the values of terms from their initial value to some other
value. \namecite{Melamed04a} presented a different
formulation, where the values of terms are initially unknown.  The
present formulation is cleaner because it obviates the need for term
values that are not semiring values (see Section~\ref{sec:sring}).

{\bf Inference rules} describe the parser's possible transitions from
one state to another.  We shall express inference rules as sequents:
$\frac{x_1 , \ldots , x_k}{y}$ means that the value of the {\bf
consequent}~$y$ depends on the values of the {\bf antecedents}~$x_1 ,
\ldots , x_k$.  For example, if we are dealing with probabilities,
then the probability of the consequent might be defined as the product
of the probabilities of the antecedents.  The exact relationship
between these values depends on the semiring, explained below.  Every
change in term value corresponds to the invocation of an inference
rule where that term is a consequent.

\begin{table*}[tb]
\tcaption{Logic~D1C.  
$w_i$ are input words; $X, Y$ and $Z$ are
nonterminal labels; $t$ is a terminal; $i$ and $j$ are word boundary
positions; $n$ is the length of the input.
\label{logic-d1c} }
\begin{centering}
\begin{tabular}{|r|c|} \hline
\multicolumn{1}{|l|}{{\bf Term Types}} & \\
terminal items
&
$\la i, t \ra$
\\
& \\
nonterminal items
&      $\left[ X; (i, j) \right]$ \\ 
& \\ 
terminating productions &
\(
X
\der
t
\)
\\
& \\
nonterminating productions & 
\(
X \der Y \; Z
\)
\\
& \\
\hline
\multicolumn{1}{|l|}{{\bf Axioms}} & \\
input words &
$\la i, w_{i} \ra$ 
for $
1 \leq i \leq n
$
\\
& \\
grammar terms & as given by the grammar 
\\
\hline
\multicolumn{1}{|l|}{{\bf Inference Rule Types}}  & \\
& \\
{\em Scan}
& \Large 
\(
\frac{
\la i, t \ra
\mbox{\Large \ ,\ }
X 
\der
t
}{
\left[
X
; 
(i-1, i)
\right]
}
\)
\\
& \\
{\em Compose} &
\Large
\(
\nonumber
\frac{
\left[ Y ; (i, j) \right]
{\Large \ ,\ }
\left[ Z ; (j, k) \right]
{\Large \ ,\ }
X \der Y \; Z
}{
\left[ X ; (i, k) \right]
}
\) \\
\ignore{
& \\ \hline
& \\
\multicolumn{1}{|l|}{\bf Goal}
&    $\left[ S; (0, n) \right]$ \\
}
& \\ \hline
\end{tabular}
\end{centering}
\end{table*}

For example, consider Logic~D1C, shown in Table~\ref{logic-d1c}.  This
is a logic for parsing under context-free grammars (CFGs) in Chomsky
Normal Form (CNF).  This logic has four term types.  Two term types
represent production rules in the grammar.  The other two term types
are items.  Each of the logic's terminal items relates a terminal
symbol to a word position.  Each of the logic's nonterminal items
relates a nonterminal symbol to a span.  Each span consists of
boundaries $i$ and $j$, which range over positions between and around
the words in the input.  The position to the left of the first word is
zero, and the position to the right of the $j$th word is $j$.  Thus,
$0 \leq i < j \leq n$, where $n$ is the length of the input.

Parser~D1C is any inference procedure based on Logic~D1C.  For every
run, Parser~D1C is initialized with axioms that represent its
grammar's production rules and the words in its input.  It then
commences to fire inferences.  A {\em Scan} inference can fire for the
$i$th word $w_{i}$ if that word appears on the right-hand side (RHS)
of a terminating production in the grammar. If a word appears on the
RHS of multiple productions (with different left-hand sides), then
multiple {\em Scan} inferences can fire for that word.  The span of
each item inferred by a {\em Scan} inference always has the form
$(i-1, i)$ because such items always span one word, so the distance
between the item's boundaries is always one.

Parser~D1C spends most of its time composing pairs of items into
larger items. It can {\em Compose} two items whenever they satisfy
both of the following constraints:
\begin{itemize}
\item {\em Immediate Dominance (ID)}. Their labels must match the
nonterminals on the RHS of a nonterminating production rule in the
grammar.
\item {\em Linear Precedence (LP).} The order of the items' spans over
the input must match the order of their labels in the antecedent
production rule.
\end{itemize}
If two spans overlap, then their order is undefined, and they cannot
{\em Compose}.  Thus, the LP constraint ensures that no part of the
input is covered more than once, so that every partial parse is a tree
(rather than a more general graph).  The reason items store their
spans it to help the parser to enforce the LP constraint.

Some previous publications included a goal item in the logic
specification.  For example, \cite{Goodman98}'s parsing logics specify
the goal of finding a constituent that covers the input text and has
the grammar's start symbol as its label.  More generally, however,
goal items can vary independently of the logic.  For example, we might
want to use Logic~D1C to find all the noun phrases in the input,
rather than a single parse for the whole sentence.  For this reason,
our logics do not specify goals.

\subsection{Search strategies}

A parsing logic specifies how terms can be inferred, but it does not
specify the order of inferences.  When a parser needs an inference to
evaluate, it consults its {\bf search strategy}. \namecite{Goodman98}
required one particular search strategy for his abstract parsing
algorithm, which depended on a topological sort of all possible terms.
\namecite{Melamed04a} used inference rules to specify a partial
order on the computations of term values, although he allowed the
order to be determined on the fly.  Here we leave all ordering
decisions to the search strategy, which may or may not consult the
logic.
A variety of search strategies are in common use.  For example, the
CKY algorithm \cite{Kasami65,Younger67} always infers smaller items
before larger ones.  Alternatively, given term costs such as negative
log-probabilities, we can run the parser as a uniform-cost search,
inferring less costly consequents before more costly ones.  If we are
interested in just the single best parse or the $n$-best parses, then
A* strategies of varying sophistication can be employed to speed up
the search \cite{KleinManning03}.  The benefit of a separate search
strategy is the usual benefit of abstraction: analyses of logics
unobscured by search strategies are applicable to a larger class of
algorithms, as we shall show in Section~\ref{sec:paramest}.

\subsection{Semirings for Parsing}

\label{sec:sring}

A semiring consists of a set, binary operators $\oplus$ and $\otimes$
over that set, and an identity element in the set for each of the two
operators.  For example, we can define a semiring over the set of
integers, where $\oplus$ and $\otimes$ are the usual addition and
multiplication operators, and the identity elements are 0 and 1.  A
semiring's set need not consist of numbers and its operators need not
be arithmetic.  

Of particular relevance here are semirings that have been proposed
specifically for the purpose of describing parsing algorithms in a
compact way.  Parsing semirings interact with parsing logics according
to the following equation:
\begin{equation}
\label{insideval0}
V(y) = \bigoplus_{\ba{c} x_1, \ldots, x_k \\
\mbox{such that } \frac{x_1, \ldots, x_k}{y} \ea} 
\bigotimes_{i=1}^k V(y_i)
\end{equation}
In this equation, $x$ and $y$ range over terms, and $V()$ is a
function that maps terms to semiring values (i.e.\ elements of the
semiring's set). The equation says that the semiring value of any term
is a sum, over all inferences where that term is a consequent, of the
product of the values of the antecedents of the inference.  The
definitions of sum and product here depend on the semiring. Some
examples will help to make these abstract ideas more concrete.

The boolean semiring over the set \{TRUE, FALSE\} defines $\oplus$ as
disjunction and $\otimes$ as conjunction.  Under this semiring, the
default term value is FALSE.  A
 term can become TRUE in one of only two ways:
\be
\item A term is TRUE if it is an axiom.  This is usually the case for
grammar terms and items representing input words. 
\item According to Equation~\ref{insideval0}, a term is TRUE if it is
the consequent of some inference rule where all the antecedents are TRUE.
\ee
If neither of the above conditions holds, then the term retains its
default FALSE value.  Starting from the parser's initial state, we can
run the parser under a Boolean semiring to determine the truth value
of the item that spans the input and has the grammar's start symbol as
its label.  A TRUE value indicates that the grammar can generate that
input.

If the grammar guiding the parser is probabilistic, then it's possible
to use an inside-probability semiring, where $\oplus$ is real addition
and $\otimes$ is real multiplication.  Under this semiring, the
grammar assigns probabilities to the grammar terms.  We can run the
parser under the inside-probability semiring to compute the total
probability of any item.  The probability of the item that spans the
input and has the grammar's start symbol as its label is the
probability of the grammar generating the input.

\namecite{Goodman98} studied the above semirings and a
variety of other semirings that are useful for parsing, including:
\bitem
\item the Viterbi semiring for computing the probability of the
single most probable derivation;
\item the Viterbi-derivation semiring for computing the single most
probable derivation;\footnote{Or the set of most probable derivations,
if there are ties.} 
\item the Viterbi-$n$-best semiring for computing the $n$ most
probable derivations; 
\item the derivation-forest semiring for computing all possible
derivations;
\item the counting semiring for computing the number of possible
derivations.
\eitem
The probabilistic semirings can be straightforwardly extended
to unnormalized weights.  The expectation semiring \cite{Eisner02} can
be used to compute expected probabilities, as well as expected feature
counts for maximum entropy models and derivatives for gradient-based
optimization methods.  All of these parsing semirings apply equally
well to all the classes of algorithms that we discuss in this article.

\subsection{Termination Conditions}

Different termination conditions are appropriate for different parsing
applications.  Most applications involve a goal item, such as an item
that spans the input and is labeled with the start symbol of the
grammar.  Then the termination condition is that no further inference
can change the value of the goal item.
Some applications, such as those in Sections~\ref{sec:w2w}
and~\ref{sec:paramest}, involve multiple goal items.  There the
termination condition must hold conjunctively for all goal items.

In practice, termination conditions often cannot be expressed solely
in terms of goal items and their values.  For example, Earley parsing
logic \cite[Section~2.1.1]{Goodman98} might be used to compute the
probability of a string under the inside semiring and a PCFG that
is not in CNF.  If the PCFG has cycles of unary productions (like $\{
A \rightarrow B, B \rightarrow A \}$), then the parser will not
terminate, because it will be computing an infinite sum.  There are
methods for computing such sums in closed form \cite{Stolcke95}, and
it is possible to augment the parser with those methods.  However,
most parsing applications resort to approximations, because such
approximations are easier to implement.  A typical implementation
limits the computing resources that a run of the parser can expend.
So, in addition to goal items, the termination condition might test
the elapsed time, the size of allocated memory, or the number of
inferences fired, possibly as a function of the input size.

\subsection{Abstract Parsing Algorithm}
\label{sec:apa}

\namecite{Goodman98} presented an abstract parsing algorithm whose
parameters are a logic, a grammar, and a semiring.  His algorithm
employs a search strategy that depends on {\em a priori} computation
of dependencies among all possible terms, so that the terms can be
topologically sorted into ``buckets.''  The parser's inferences are
then fired in the order of their consequents' buckets.  Goodman's
algorithm also assumes that the termination condition is based on a
particular goal item.  Table~\ref{napa} presents a more detailed and
more general abstract parsing algorithm, where the search strategy and
termination condition are parameters, along with the logic, the
grammar, the semiring, and the input text.

The parser initializes all possible terms with $0_R$, the value of the
additive identity element of the semiring.\footnote{A typical
implementation would not store terms that have this value, so this
step would do nothing.}  It then re-initializes axiom terms. It
consults the grammar for the value $G(p)$ of each grammar term $p'$.
The grammar must be compatible with the semiring so that $G(p)$ is
always a semiring value (i.e.\ an element of the semiring's set).
Ordinarily, all other axioms are assigned the semiring's
multiplicative identity element $1_R$. However, if the input is
nondeterministic, then the input axioms can take other values.  E.g.,
they might be weighted by the acoustic module of a speech recognizer.

After initialization, the parser enters its main loop.  On each
iteration of the main loop, the parser first calls the search strategy
to select a set of antecedent terms.  The parser places no
restrictions on how the search strategy might do so.  However, a
typical search strategy would keep track of which sets of antecedents
it selected previously, to avoid duplication of effort.  It would then
return antecedent sets that have either (a) never been selected
before, or (b) have had one of their element's values changed since
the previous time they were selected. If the search strategy cannot
find a set of antecedents with one of these properties, it would
return the empty set, which might satisfy the termination condition.

When the parser receives the set of antecedents from the search
strategy, it passes them to the logic.  The logic compares the
antecedents to the signatures of its inference rules.  For every
matching inference rule, the logic instantiates every possible
consequent.  It passes all the consequents from all matching inference
rules back to the parser.

\begin{table}
\caption{Abstract Parsing Algorithm \label{napa}}
Input: 
\bitem
\item logic $L$, 
\item grammar $G$, 
\item semiring $R$, 
\item search strategy $S$,
\item termination condition $C$, 
\item text $T$ 
\eitem
\begin{algorithmic}[1]
\ForAll{possible terms $x \in L$}
	\State	$V(x) = 0_R$
		\Comment{$0_R$ is the additive identity element of $R$.}
\EndFor
\ForAll{axioms $p' \in L$ corresponding to a production rule $p \in G$}
\State	$V(p') = G(p)$	\Comment{$G(p)$ is the value that $G$ assigns to $p$.}
\EndFor
\ForAll{axioms $w' \in L$ corresponding to word $w \in T$}
\State	$V(w') = T(w)$	\Comment{$T(w) = 1_R$ if $T$ is unambiguous.} 
\EndFor
\ForAll{other axioms $a \in L$}
	\State	$V(a) = 1_R$
		\Comment{$1_R$ is the multiplicative identity element of $R$.}
\EndFor
\Repeat
\State	get a set of antecedents $X = \{x_1, \ldots, x_k\}$ from $S$

\ForAll{inferences $I \in L$ such that $X$ unifies with the antecedents of $I$}
	\ForAll{possible terms $y$ that unify with the conseq. of ``$I$ unified with $X$''}
		\State  
		\begin{equation}
		\label{sas}
		SetAntSet(y) = SetAntSet(y) \cup \{X\}
		\end{equation}
		\State
		\begin{equation}
		\label{insideval}
		V(y) = \bigoplus_{X \in SetAntSet(y)} \bigotimes_{i = 1}^{|X|} V(x_i)
		\end{equation}
	\EndFor
\EndFor
\Until{$C$ satisfied}
\ignore{
\ForAll{ goal terms $y \in L$}
\State Output $y$ with $V(y)$.
\EndFor
}
\end{algorithmic}
\hrule
\end{table}

The parser then performs two updates for each consequent.
Equation~\ref{sas} updates the record of the consequent's set of
antecedent sets (SetAntSets).  Equation~\ref{insideval} uses the two
operators of the semiring to recompute the consequent's value from the
values of its SetAntSets. The SetAntSets data structure is the same as
the set of ``back-pointers'' necessary to represent a packed
derivation forest computed under the derivation-forest semiring.  For
some other semirings, this structure is partially or totally
redundant, and more efficient updates are possible.  For example,
under the Viterbi-derivation semiring, it is necessary to keep track
of only the highest-probability antecedent set.  However, we don't
know of any way to optimize Equation~\ref{sas} that would be correct
for {\em all} parsing semirings.

Under some semirings, Equation~\ref{insideval} can also be
optimized to reuse previous computations, along the lines of
\cite{Goodman98}'s ``Item Value Formula.''  However, such
optimizations can invalidate nonmonotonic parsing logics, such as
those that involve pruning (see Section~\ref{sec:effic}).  Another
possible optimization is to move Equation~\ref{insideval} outside the
main loop, and compute $V(y)$ just once for each $y$.  This
optimization is impractical in the common scenario where the
termination condition involves a time limit.  Of all the abstract
parsing algorithms that we are aware of, the algorithm in
Table~\ref{napa} is the only one that admits all known parsing logics,
semirings, search strategies, and termination conditions.

\section{Generalized Parsers}
\label{sec:mp}

In an ordinary parser, the input is a single string, and the grammar
ranges over strings.  A convenient way for an SMT system to create and
use tree-structured translation models is via generalizations of
ordinary parsing algorithms that allow the input to consist of string
tuples and/or the grammar to range over string tuples.  The kind of
string tuples that are most relevant here are texts that are
translations of each other, also called {\bf parallel texts} or {\bf
multitexts}. Each multitext consists of {\bf component texts} or {\bf
components}.  Borrowing from vector algebra, we shall use {\bf
dimension} as a synonym for {\bf component}, so the number of
components in a given multitext is its {\bf dimensionality}.

\begin{figure}
\centerline{
(a)
\psfig{figure=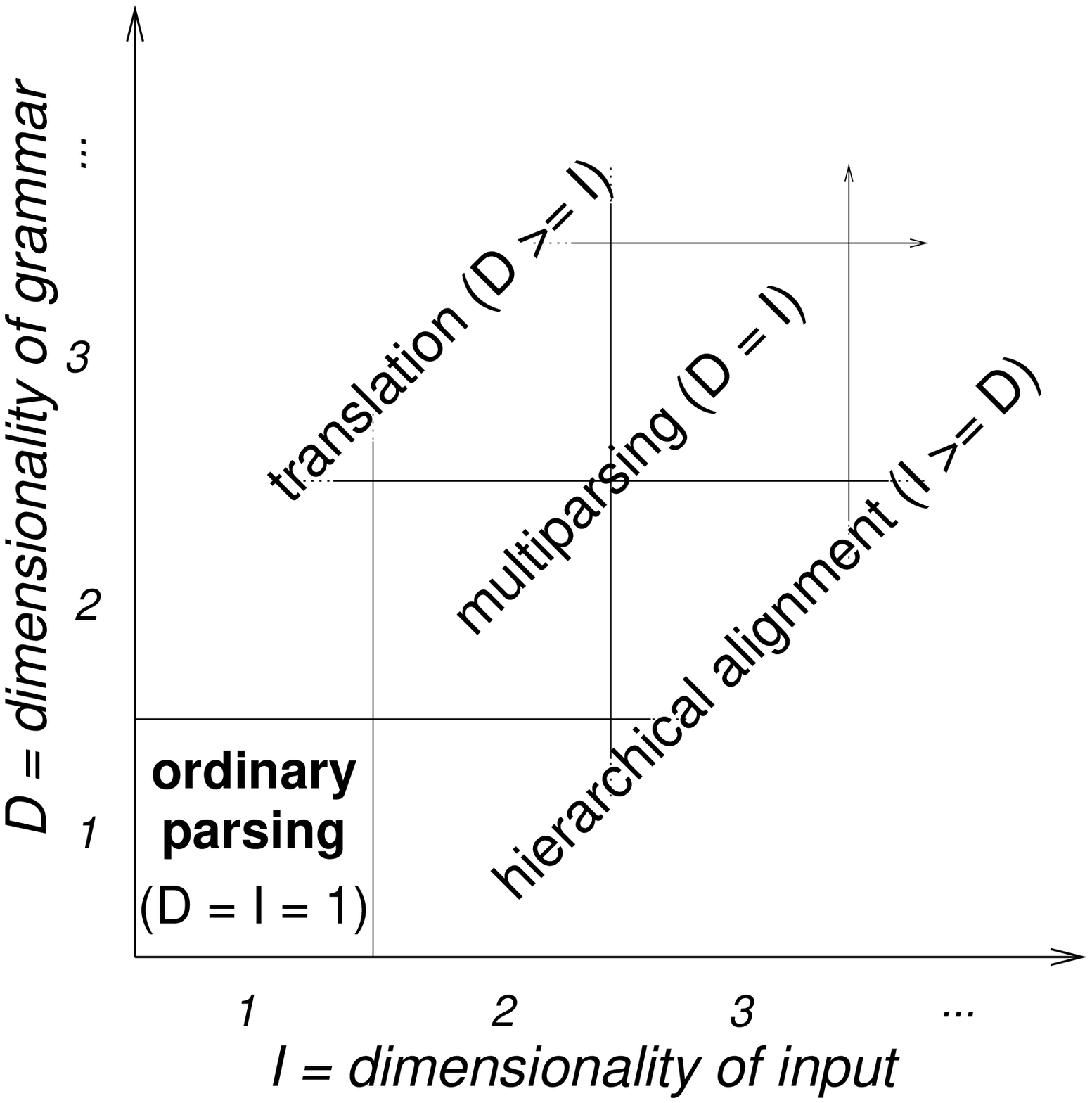,height=2.75in}
\hfill
(b) 
\psfig{figure=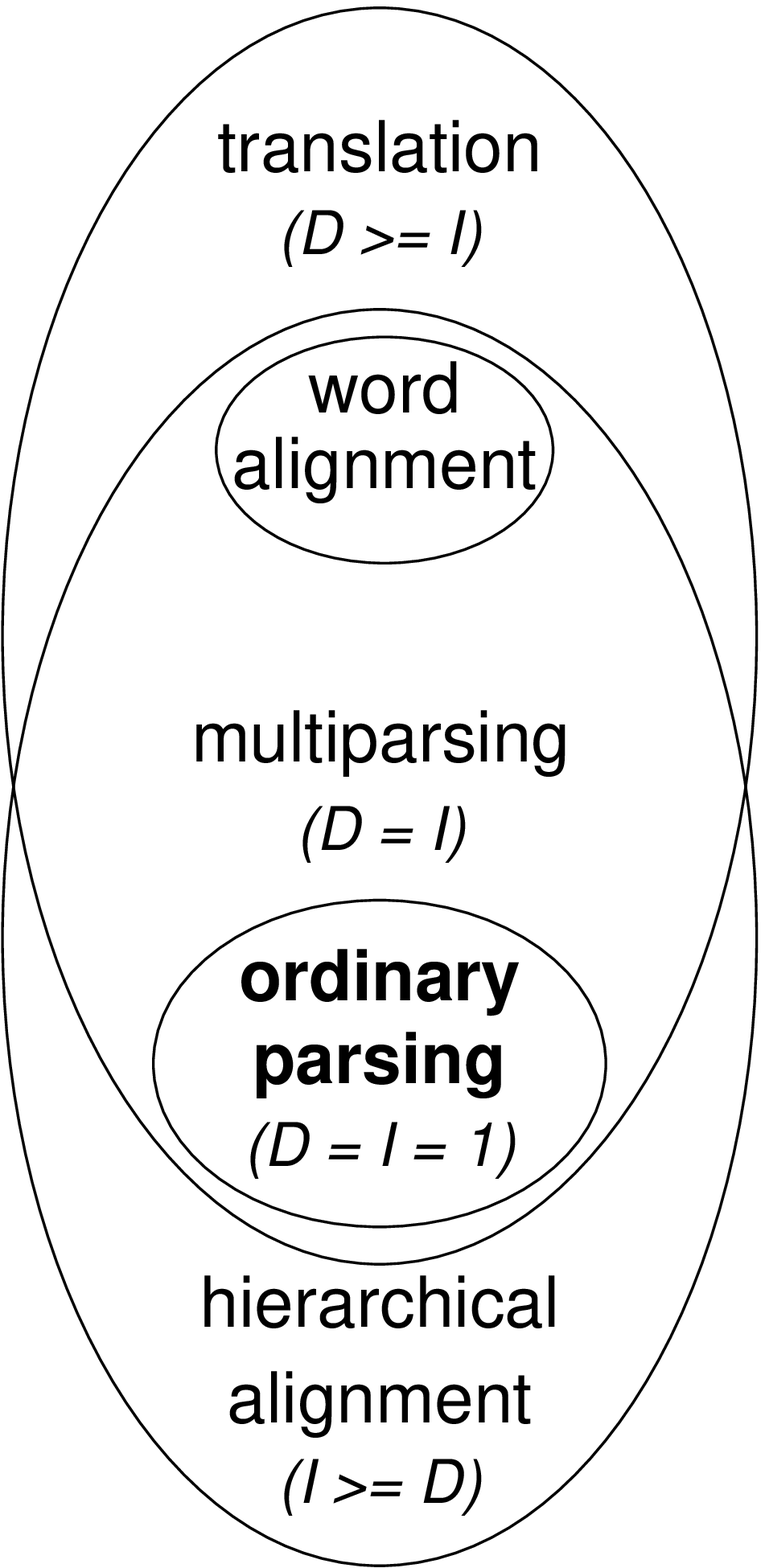,height=2.75in}
}
\caption{Two perspectives on the space of generalized parsers.
\label{venn}
\label{parsematrix}}
\end{figure}

Figure~\ref{parsematrix} shows some of the ways in which ordinary
parsing can be generalized.  A {\bf multiparser}\footnote{An
equivalent term is {\bf synchronous parser} \cite{Melamed03}.} is an
algorithm that can infer the tree structure of each component text in
a multitext and simultaneously infer the correspondence relation
between these structures.\footnote{A suitable set of monolingual
parsers can also infer the tree structure of each component, but it
cannot infer the correspondence relation between these structures.} 
When a parser's input can have fewer dimensions than the parser's
grammar, we call it a {\bf translator}.  When a parser's grammar can
have fewer dimensions than the parser's input, we call it a {\bf
hierarchical aligner}, or just an {\bf aligner} when the context is
unambiguous.
\footnote{This class of algorithms has many names
in the literature, such as structural matching
\cite{Matsumoto+93}, sub-sentential alignment \cite{Groves+03}, and
synchronization \cite{Melamed04a}.  There are also names for proper
subclasses of algorithms, such as ``tree alignment'' \cite{Meyers+96} and
``biparsing alignment'' \cite{Wu00}.  Although ``alignment''
traditionally referred to monotonic relations (i.e.\ without crossing
correspondences), we follow what seems to have become standard
nomenclature in computational linguistics.} 
The corresponding processes are called {\bf
multiparsing, translation} and {\bf hierarchical alignment},
respectively.

Many previously published algorithms can also be viewed as generalized
parsers \cite[e.g.]{AhoUllman69,Wu96,Alshawi96,Hwa+02}.  Some of these
other parsers are fundamentally similar to our parsers and to each
other.  Others are superficially similar but subtly different.  For
example, some of the algorithms that have been put forth as
generalizations of the CKY algorithm turn out to be more complicated
than our generalization (see Section~\ref{sec:logicc}), and therefore
possibly more complicated than necessary.  As we shall show, the
similarities and differences are easier to see when the semiring, the
search strategy, and the terminating conditions are abstracted away.

Taking advantage of the
clarity provided by these abstractions, we shall elucidate the
relationships between several classes of generalized parsers:
\bitem
\item The class
of ordinary parsers is a proper subclass of the class of multiparsers,
because the grammars and logics used for ordinary parsing are special
cases of the grammars and logics used for multiparsing.  
\item The class of multiparsers is a proper subclass of the class of
translators, because the logics of multiparsing are a subset
of the logics of translation.  
\item The class of multiparsers is
also a proper subclass of the class of hierarchical aligners, because
the grammars used for multiparsing are a subset of the grammars used
for hierarchical alignment.  
\eitem These relationships could not have been
spelled out as precisely without the abstractions in
Section~\ref{sec:anat}.

Most of the rest of this article is a guided tour of the generalized
parsers that are useful for SMT by Parsing.  The next section
describes the kind of grammar that generalized parsers use, and
presents a particular grammar formalism that will serve as a vehicle
for our tour.  The three sections after the next give detailed
examples of multiparsers, translators, and hierarchical aligners.
Then, Sections~\ref{sec:w2w} and~\ref{sec:paramest} present two
additional generalized parsers that are necessary for a complete
system.  All of the algorithms on the tour are special cases of the
abstract parsing algorithm in Table~\ref{napa}.

\section{Grammars for Generalized Parsing}
\label{sec:mtg}

To parse string tuples, we need grammars that can evaluate structures
over string tuples, rather than just structures over strings.
Grammars that can evaluate structures over string tuples are often
called {\bf synchronous grammars}.\footnote{They were originally
called {\bf transduction grammars} \cite{AhoUllman69}, but we follow
the majority of the literature in avoiding this term so as to
de-emphasize the input-output connotation of ``transduction''
(cf. \namecite[p.\ 378]{Wu97}).}  This article is not about grammar formalisms,
but for expository purposes it is convenient to use one particular
formalism as a running example.  Our choice of synchronous grammar
formalism is informed by certain properties of popular monolingual
grammar formalisms and parsing algorithms.  To minimize the conceptual
leap from ordinary parsing to generalized parsing, we shall employ a
grammar formalism that is similar to CFG, in that it uses production
rules and is associated with a context-free derivation process.

Another consideration is that popular monolingual grammar formalisms,
such as CFG, TAG, and CCG explicitly express subcategorization frames,
and recognize that subcategorization frames often have more than two
dependents.  In formalisms that involve production rules, such
subcategorization frames are expressed via production rules that have
more than two nonterminals on the RHS.  For inferring such
productions, it is always more efficient to binarize the grammar
(either explicitly or implicitly) than to allow a parser to compose
more than two parts of the input at a time.  However, in general,
binarization of synchronous production rules can result in
discontinuous nonterminals.

\begin{figure}
\centerline{
\(
\begin{array}{c} S[went] \\  S[pashol] \end{array}
\!\! \der \!\!
\begin{array}{c} 
NP[Pat]^1 \; V[went]^2 \; Adv[home]^3 \; Adv[early]^4 \\ 
Adv[damoy]^3 \; NP[Pat]^1 \; Adv[rano]^4 \; V[pashol]^2
\end{array}
\)
\hfill
\raisebox{-.5in}{\psfig{figure=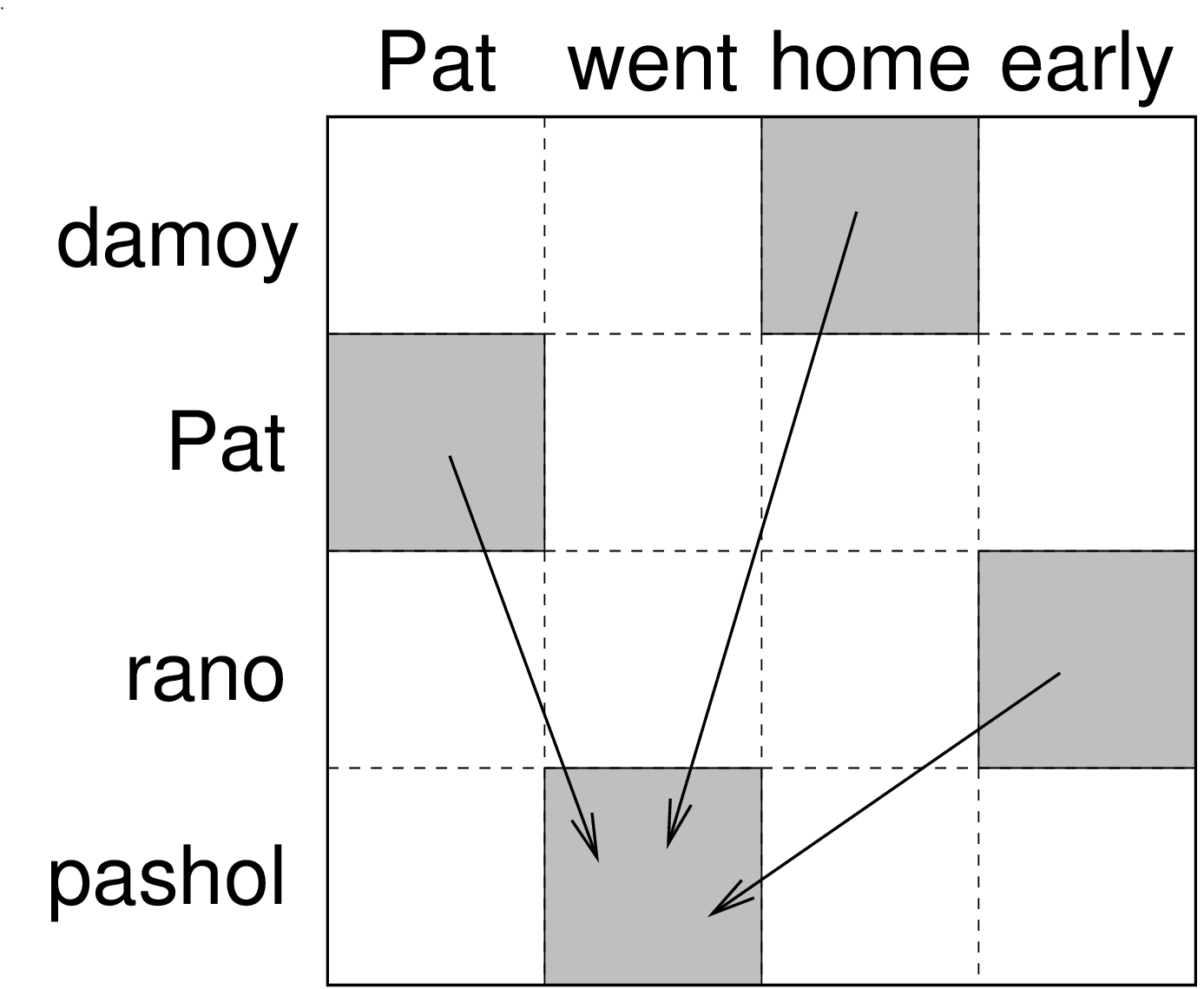,height=30mm}}
}
\caption{
Discontinuous constituents are practically unavoidable in
generalized parsing, for example when inferring this production rule in
English and (transliterated) Russian.  In the illustration on the
right, arrows are 2D bilexical dependencies, and shaded squares are 2D
constituents.
\label{fig:estlinks}}
\end{figure}

Early synchronous grammars, such as syntax-directed transduction
schemata (SDTSs) \cite{AhoUllman69} and their subclass of inversion
transduction grammars (ITGs) \cite{Wu97}, were defined for contiguous
constituents only.  Therefore, these formalisms cannot generate
certain important multitext correspondence patterns using binary
derivation trees.
For example, an SDTS that allows up to four symbols on the RHS of a
production rule can generate the correspondence pattern in
Figure~\ref{fig:estlinks}, but no SDTS with a lower limit can generate
it.
The distinguishing characteristic of such patterns is that
no two constituents are adjacent in more than one dimension.
Each set of sibling constituents in such a pattern must be
encapsulated in the RHS of a single nonterminating production rule,
like the one shown in Figure~\ref{fig:estlinks}.  

\begin{figure}
\raisebox{-.5in}{\psfig{figure=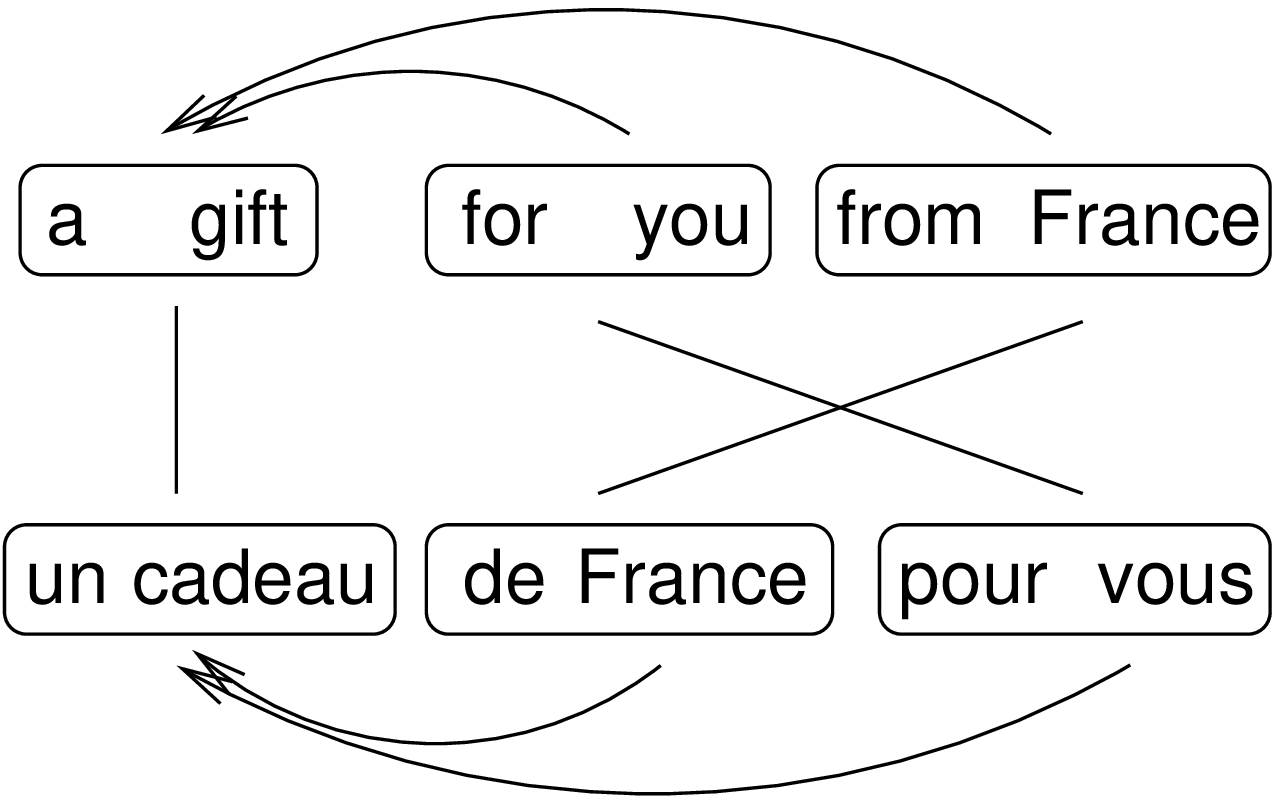,height=35mm}}
  \hfill 
\raisebox{-.5in}{\psfig{figure=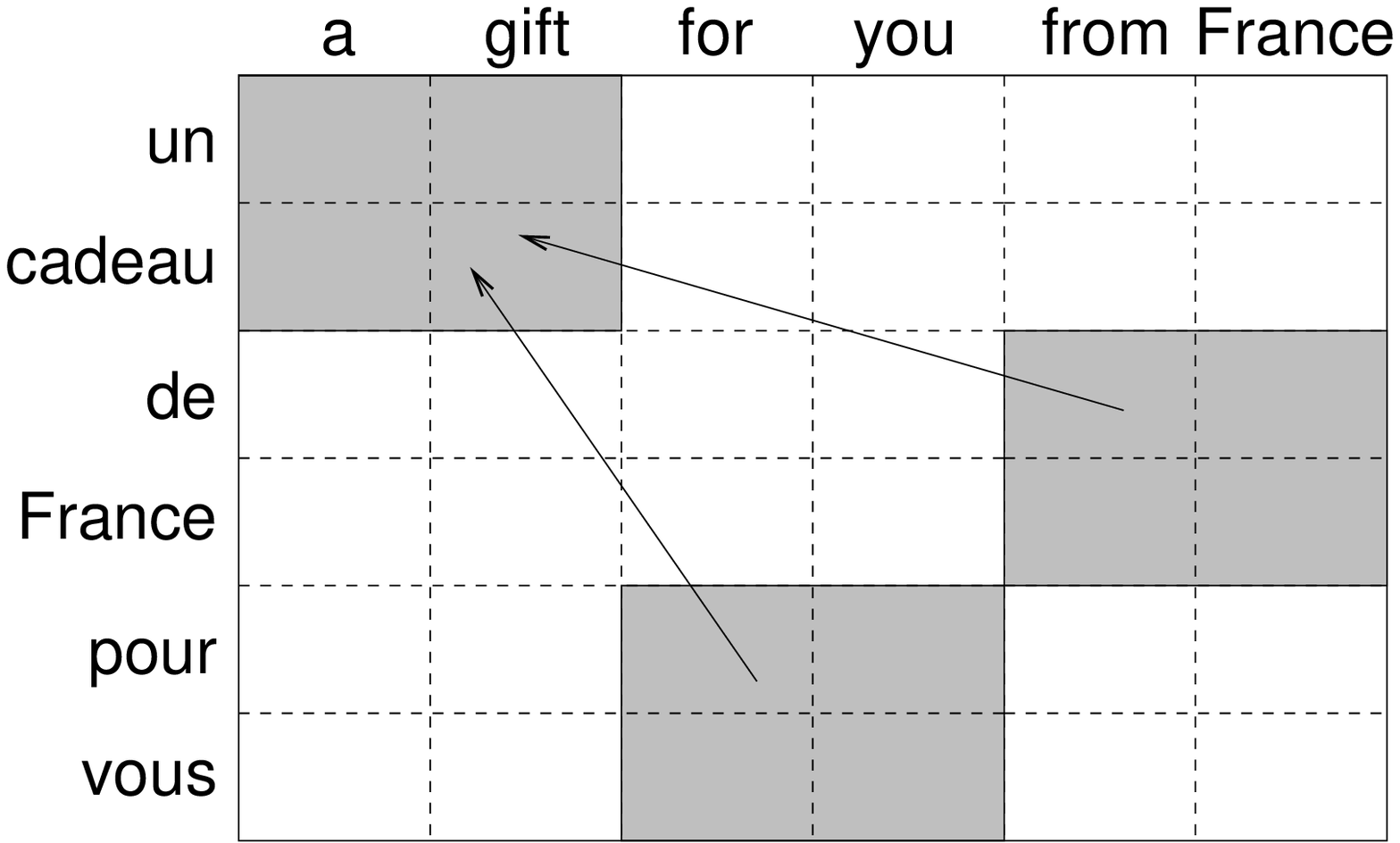,height=35mm}}
\caption{
Discontinuous constituents are required for bilexical
parsing, even for simple bitexts in English and French.  
 \label{discconsts}}
\end{figure}

If the grammar is bilexical, then even productions with only three
nonterminals on the RHS can require discontinuous constituents for
binarization.  This case can arise, for example, when two
prepositional phrases switch places in translation, as shown in
Figure~\ref{discconsts}.
Bilexical parsers typically compose dependents with the head-child,
rather than with other dependents, because otherwise some items would
need to keep track of multiple lexical heads, increasing computational
complexity.  Each of the dependents in Figure~\ref{discconsts} is
adjacent to the head-child in only one dimension.  Regardless of which
dependent is attached first, a discontinuous item will result.

Synchronous grammar formalisms that do not allow discontinuous
constituents are unlikely to have adequate coverage, even for multitexts
involving languages that are syntactically similar \cite{ZensNey03}.
\namecite{Simard+05} have presented empirical evidence that an SMT
system can perform better if it can manipulate discontinuous
constituents.  To our knowledge, the simplest synchronous grammar
formalism that deals in discontinuities is generalized multitext
grammar (GMTG)
\cite{Melamed+04}.  
GMTGs are a complete generalization of
CFGs to the synchronous case: GMTGs can express arbitrary tree
structures over arbitrarily many parallel texts.  This article uses
GMTG as a running example of a tree-structured translation model
because GMTG is the simplest formalism that can be used to illustrate
the concepts that we consider important.  More sophisticated
formalisms would be necessary to represent a variety of translational
divergence patterns \cite{Dorr94}, but our abstract parsing algorithm
can accommodate them without modification.  In a similar spirit,
\namecite[Section~2-B.2]{Goodman98} presented a parsing logic for
tree-adjoining grammars that can be used in his abstract parsing
algorithm without modification.

Every GMTG is a $D$-GMTG for some integer constant~$D>0$, and it
generates multitexts with $D$ components.  Thus, 1-GMTGs generate
ordinary texts and 2-GMTGs generate bitexts.  A GMTG has disjoint sets
of terminals $T$ and nonterminals $N$.
We often group terminals and nonterminals into vectors that we call
{\bf links}.
Links express the translational equivalence between their components.
In GMTG applications, the different components of a link will often
come from largely disjoint subsets of $T$ or $N$, representing the
vocabularies and linguistic categories of different languages. Every
link generated by a $D$-GMTG has $D$ components, some of which may be
inactive\footnote{Inactive component are distinct from components that
contain the empty string \cite{Melamed+04}.  This distinction obviates
the need to keep track of the positions of empty strings during
parsing.}.  An inactive link component indicates that the active
components vanish in translation to the inactive component.

Each GMTG also has a set of {\bf production rules} (or just {\bf productions}
for short).  A production in 2-GMTG might look like this:
\begin{equation}
\label{depprod-eg}
\left[
\begin{array}{@{}c@{}} (X) \\  (Y) \end{array}
\right]
\der 
\left[
\begin{array}{@{}c@{}} (A^1 \; e \; C^3) \\ (A^2 \; D^1) \end{array}
\right]
\end{equation}
There is one row per production component, on both the left-hand side
(LHS) and the right-hand side (RHS).  Each symbol in parentheses on
the LHS is a nonterminal.  
Each component of the RHS is a string of terminals and/or indexed
nonterminals. The indexes are not part of the nonterminal labels; they
exist only in production rules.  The same nonterminal symbol may
appear multiple times in the same component or in different
components, either with the same index or with a different index (like
A).

The indexes express translational equivalence: All the nonterminals
with the same index constitute a link.  The derivation process
rewrites linked nonterminals as atomic units.  Some nonterminals on
the RHS might have no translation in some components, in which case
there will be no co-indexed nonterminal in those components (as for
$A^2$ and $C^3$).  The derivation process rewrites such nonterminal
links like any other link, generating parse subtrees that are inactive
in some components.  In the limit, a nonterminal symbol in one
dimension might not be coindexed with any other nonterminal symbol in
its production rule.  Repeated rewriting of such degenerate
nonterminal links can generate arbitrarily deep one-dimensional
subtrees that correspond to other dimensions only at their root.  A
GMTG can generate tuples of such subtrees to represent translational
equivalence among ``phrases,'' a concept that is currently popular in
SMT \cite[e.g.]{Koehn+03}.  Terminals never have indexes because they
are never rewritten.

The production rule notation described above, which is the original
notation of \namecite{Melamed+04}, uses superscripts to
superimpose information about translational equivalence on top of
information about the linear order of constituents.  This notation
highlights the relationship between GMTG and the familiar CFG.
Unfortunately, this notation is not conducive to describing the way
that grammar terms interact with the inference rules in parsing
logics.  In order to specify inference rule signatures completely and
compactly, we introduce an alternative notation for GMTG productions
that have nonterminals on the RHS.  The new notation separates
information about translational equivalence from information about the
linear order of constituents, enabling independent reference to each
type of information.

Here is Production~(\ref{depprod-eg}) rewritten in the new notation:
\begin{equation}
\label{depprod-eg-alt}
\begin{array}{@{}c@{}} X \\  Y \end{array}
\der \Join
\left[
\begin{array}{@{}c@{}} [1, 2, 3] \\ {[4, 1]} \end{array}
\right]
\left(
\begin{array}{@{}c} \text{A } \\ \text{D } \end{array}
\begin{array}{c} \text{ e } \\ \emptyset \end{array}
\begin{array}{c} \text{ C } \\ \emptyset \end{array}
\begin{array}{c} \emptyset \\ \text{ A} \end{array}
\right) .
\end{equation}
In this notation, nonterminal links are written in columns, and their
linear order is indicated by a preceding vector of special data
structures called {\bf precedence arrays}, one array per component.
E.g., the precedence array in the second component above is $[4,1]$.
The first index in this array is 4, referring to the fourth column in
the link vector, and indicating that $A$ comes first in that
component.  The special symbol $\emptyset$ acts as a placeholder for
inactive link components.  The indexes in a precedence array never
refer to links that are inactive in their component.  If the LHS link
is inactive in a given component, then all the links on the RHS must
also be inactive, and vice versa.  In that case, the component is
called {\bf inactive} and the precedence array must be empty.
Precedence arrays are more informative than the role templates used by
\namecite{Melamed03}, because role templates obscure link information.
The $\Join$ (``join'') operator rearranges the symbols in each
component's link vector according to that component's precedence array
to recover the original production rule notation.  For example,
\begin{equation}
\label{join-eg}
\Join
\left[
\begin{array}{@{}c@{}} 
[ 1, 2, 3 ]  \\
{[ 4, 2, 1, 5 ]} \\ 
{[3, 2, 4]} \end{array}
\right]
\left(
\begin{array}{@{}c} \text{A} \\ \text{Y} \\ \emptyset \end{array}
\begin{array}{@{}c} \text{B} \\ \text{X} \\ \text{U} \end{array}
\begin{array}{@{}c} \text{C} \\ \emptyset \\ \text{V} \end{array}
\begin{array}{@{}c} \emptyset \\ \text{W} \\ \text{T} \end{array}
\begin{array}{@{}c@{}} \emptyset \\ \text{Z} \\ \emptyset \end{array}
\right)
=
\begin{array}{@{}c@{}} \text{A}^1 \text{B}^2 \text{C}^3  \\ 
 \text{W}^4 \text{X}^2 \text{Y}^1 \text{Z}^5 \\
 \text{V}^3 \text{U}^2 \text{T}^4
\end{array}
\normalsize
\end{equation}
All the precedence arrays in a given production rule constitute a {\bf
precedence array vector (PAV)}.  

\begin{figure*}
\centerline{\psfig{figure=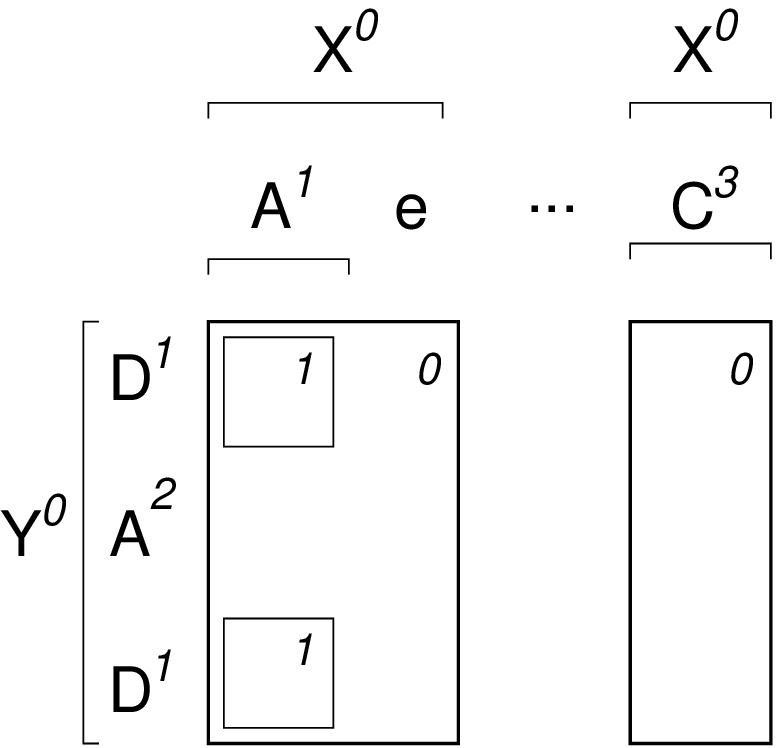,height=1.5in}}
\caption{Example of a GMTG production involving discontinuities on
both the LHS and the RHS.  Numbers indicate co-indexation.
\label{disctree}}
\end{figure*}

Precedence arrays can express discontinuities.  They can also indicate
how to arrange parts of discontinuous subconstituents.  For example,
suppose that the first component of Production~(\ref{depprod-eg-alt})
had a gap between the e and the C.  Suppose further that the D in the
second component contained a gap, and that the A in that component
filled that gap.  Then the production might be written like this:
\begin{equation}
\label{depprod-eg2-alt}
\begin{array}{@{}c@{}} X \\  Y \end{array}
\der \Join
\left[
\begin{array}{@{}c@{}} [1, 2 ; 3] \\ {[1, 4, 1]} \end{array}
\right]
\left(
\begin{array}{@{}c} \text{A} \\ \text{D} \end{array}
\begin{array}{c} \text{ e } \\ \emptyset \end{array}
\begin{array}{c} \text{C} \\ \emptyset \end{array}
\begin{array}{c} \emptyset \\ \text{A} \end{array}
\right)
\end{equation}
In the first component, the semicolon in the precedence array
indicates the position of the gap.  In the second component, the
precedence array indicates that the two parts of D (the nonterminal in
the first link) should wrap around the A (the nonterminal in the
fourth link).  The multitree fragment generated by this production
rule is illustrated in Figure~\ref{disctree}.  
Precedence arrays are more general than permutations, because
precedence arrays can refer to the same position more than once, as
for the $D$ in the second component above.

For each production in the original CFG-style notation, there are many
ways to re-express it in the new notation.  The existence of multiple
ways to express the same constraint is called spurious ambiguity, and
it leads to wasted effort during parsing. To avoid spurious ambiguity,
we stipulate a normal form for production rules in the new notation.
The normal form requires that, if the arrays in the PAV are
concatenated, then the first appearance of an index $i$ must precede
the first appearance of an index $j$ for all $i < j$, except where the
arrangement is incompatible with an earlier choice of indexes.  We
could, for example, obtain the same result in Equation~\ref{join-eg}
if we put $\emptyset Z \emptyset$ before $\emptyset W T$ and
switch their indexes in the 2nd and 3rd precedence arrays.  However,
the normal form requires the 2nd precedence array to be [4, 2, 1, 5],
not [5, 2, 1, 4], so $\emptyset Z \emptyset$ must be listed last in
the link vector.  There is a one-to-one correspondence between
production rules in the new notation that are in normal form and
production rules in the original notation.

For simplicity, we shall limit our attention to GMTGs in Generalized
Chomsky Normal Form (GCNF) \cite{Melamed+04}.
This normal form allows simpler algorithm descriptions than the normal
forms used by \namecite{Wu97} and \namecite{Melamed03}.  In GCNF,
every GMTG production is either a terminating production or a nonterminating
production.  {\bf Terminating productions} have the form
\begin{equation}
\label{yieldtemplate}
\begin{array}{@{}c@{}} \vdots \\ \emptyset \\ X \\ \emptyset \\ \vdots \end{array}
\der 
\begin{array}{@{}c@{}} \vdots \\ \emptyset \\ t \\ \emptyset \\ \vdots \end{array}
\end{equation}
All components except for one are inactive.  The active component has
a single nonterminal symbol on the LHS and a single terminal symbol on
the RHS.
{\bf Nonterminating productions} have the form
\begin{equation}
\label{deptemplate}
\begin{array}{@{}c@{}} X_1 \\ \vdots \\ X_D \end{array}
\der \Join
\left[
\begin{array}{@{}c@{}} \pi_1 \\ \vdots \\ \pi_D \end{array}
\right]
\left(
\begin{array}{@{}c@{}} Y_1 \\ \vdots \\ Y_D \end{array}
\begin{array}{@{}c@{}} Z_1 \\ \vdots \\ Z_D \end{array}
\right)
\end{equation}
where every X, Y, and Z is either a nonterminal symbol or $\emptyset$
and every $\pi$ is a precedence array.  In GCNF, every nonterminating
production must have exactly two nonterminal links on the RHS.  These
two links may or may not have any active dimensions in common.
However, whenever $X_i$ is $\emptyset$, both $Y_i$ and $Z_i$ must also
be $\emptyset$, and vice versa.  Each link can have an arbitrary
number of discontinuities, which means that the precedence arrays can
be arbitrarily long.  However, every index in those arrays is either 1
or 2.

\begin{figure*}
(a) ordinary tree view 
\raisebox{.25in}{
\Tree
[.{$ \left[ \begin{array}{@{}c@{}} S [1,2] \\ S [2,1] \end{array} \right] $}
[.{$ \left[ \begin{array}{@{}c@{}} NP \\ NP [2,1] \end{array} \right] $}
[.{$ \left[ \begin{array}{@{}c@{}}  N \\ N  \end{array} \right] $}
[.{$ \left[ \begin{array}{@{}c@{}}  PAS \\ \mbox{}  \end{array} \right] $}
{$ \left[ \begin{array}{@{}c@{}} \text{Pasudu} \\ \mbox{}  \end{array} \right] $}
]
[.{$ \left[ \begin{array}{@{}c@{}} \mbox{} \\  DISH  \end{array} \right] $}
{$ \left[ \begin{array}{@{}c@{}}  \mbox{} \\ \text{dishes}  \end{array} \right] $}
]
]
[.{$ \left[ \begin{array}{@{}c@{}} \mbox{} \\ D  \end{array} \right] $}
{$ \left[ \begin{array}{@{}c@{}} \mbox{} \\ \text{the}  \end{array} \right] $}
]
]
[.{$ \left[ \begin{array}{@{}c@{}} V \\ V   \end{array} \right] $}
[.{$ \left[ \begin{array}{@{}c@{}} MIT \\ \mbox{}  \end{array} \right] $}
{$ \left[ \begin{array}{@{}c@{}} \text{moy} \\ \mbox{} \end{array} \right] $}
]
[.{$ \left[ \begin{array}{@{}c@{}} \mbox{} \\ WASH  \end{array} \right] $}
{$ \left[ \begin{array}{@{}c@{}} \mbox{} \\ \text{Wash}  \end{array} \right] $}
]
]
]
}
\vspace*{.25in}
\newline
(b) parallel view ~\raisebox{-1.5in}{\hspace*{1in}
\psfig{figure=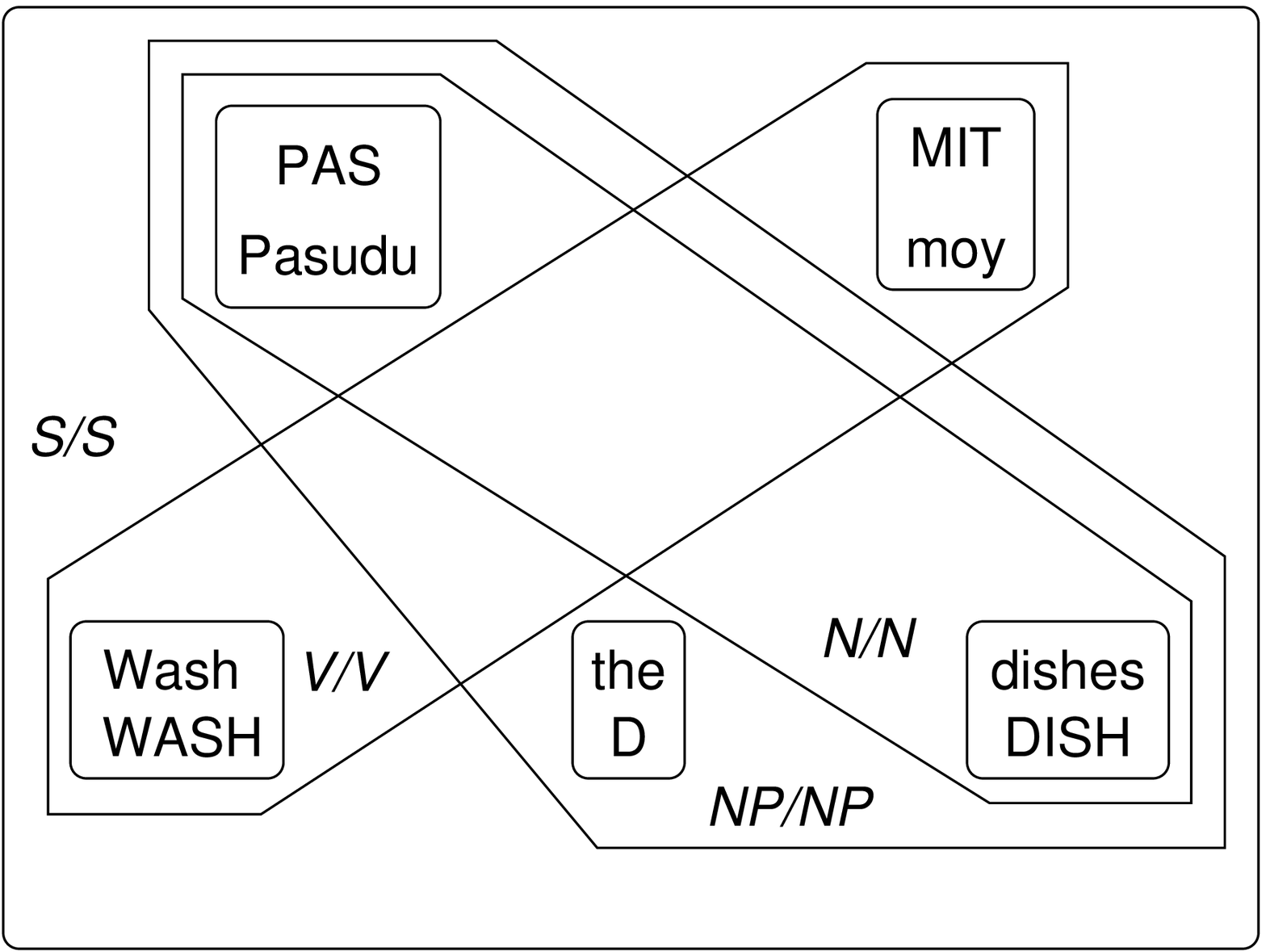,height=2in}}
\vspace*{.25in}
\newline
(c) 2D view ~\raisebox{-1.5in}{\hspace*{1.5in}
\psfig{figure=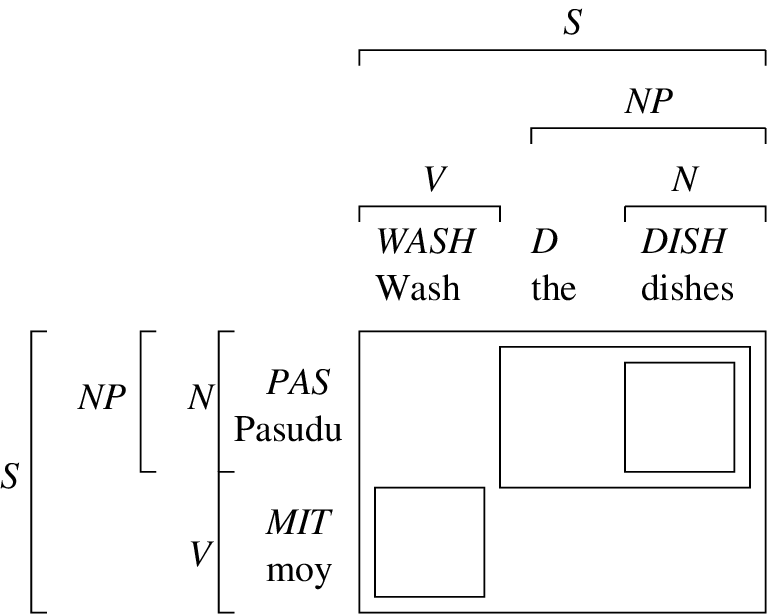,height=2in}}
\caption{A 2D multitree in English and transliterated
Russian.  The three representations are equivalent: (a) Every internal
node is annotated with the linear order of its children, in every
component where there are two children.  (b,c) Polygons are 
constituents. 
\label{mtree}}
\end{figure*}


\begin{table}
\tcaption{A 2-GMTG with the production rules in (a) can derive the
multitree in Figure~\ref{mtree} as shown in (b).
Production~(\ref{cleanprod}) is not used in the derivation.
\label{mtg-eg}}
(a)
\begin{eqnarray}
\begin{array}{@{}c@{}}
 S \\  S 
\end{array} 
 & \!\!\! \der \!\!\! &
\Join \left[ \ba{@{}c@{}} {[1 , 2]} \\ {[2 , 1]} \ea \right]
\left(
\begin{array}{@{}c@{}}
NP \\ NP
\end{array}
\begin{array}{c@{}}
  V \\ V
\end{array}
\right)
\\
\begin{array}{@{}c@{}}
  NP  \\  NP  
\end{array}
 & \!\!\! \der \!\!\! &
\Join \left[ \ba{@{}c@{}}  {[1]} \\  {[2 , 1]} \ea \right]
\left(
\begin{array}{@{}c@{}}
 N  \\  N 
\end{array}
\begin{array}{c@{}}
\mbox{} \\  D 
\end{array}
\right)
\\
\begin{array}{@{}c@{}}
 V  \\ V 
\end{array}
 & \!\!\! \der \!\!\! &
\Join \left[ \ba{@{}c@{}} {[1]}  \\ {[2]} \ea \right]
\left(
\begin{array}{@{}c}
 MIT \\ \mbox{} 
\end{array}
\begin{array}{@{}c@{}}
\mbox{} \\ WASH 
\end{array}
\right)
\\
\begin{array}{@{}c@{}}
  N  \\  N  
\end{array}
 & \!\!\! \der \!\!\! &
\Join \left[ \ba{@{}c@{}} {[1]} \\ {[2]} \ea \right]
\left(
\begin{array}{@{}c}
 PAS   \\  \mbox{} 
\end{array}
\begin{array}{@{}c@{}}
  \mbox{} \\  DISH  
\end{array}
\right)
\\
\label{washprod}
\begin{array}{@{}c@{}}
\emptyset  \\  WASH  
\end{array}
 & \!\!\! \der \!\!\! &
\begin{array}{@{}c}
  \emptyset  \\ \text{Wash}  
\end{array}
\\
\label{cleanprod}
\begin{array}{@{}c@{}}
  \emptyset \\ WASH  
\end{array}
 & \!\!\! \der \!\!\! &
\begin{array}{@{}c}
\emptyset \\   \text{clean}  
\end{array}
\\
\begin{array}{@{}c@{}}
  \emptyset \\ D 
\end{array}
 & \!\!\! \der \!\!\! &
\begin{array}{@{}c}
  \emptyset  \\ \text{the}  
\end{array}
\\
\begin{array}{@{}c@{}}
  \emptyset  \\ DISH  
\end{array}
 & \!\!\! \der \!\!\! &
\begin{array}{@{}c}
  \emptyset  \\ \text{dishes}  
\end{array}
\\
\begin{array}{@{}c@{}}
  PAS \\ \emptyset 
\end{array}
& \!\!\! \der \!\!\! &
\begin{array}{@{}c@{}}
 \text{Pasudu}  \\ \emptyset 
\end{array}
\\
\begin{array}{@{}c@{}}
 MIT  \\  \emptyset 
\end{array}
 & \!\!\! \der \!\!\! &
\begin{array}{@{}c}
 \text{moy} \\ \emptyset  
\end{array}
\end{eqnarray} 
(b)
\begin{eqnarray}
\begin{array}{c} S \\ S \end{array} 
& \rightarrow & \nonumber
\begin{array}{@{}c} NP^1 \\  V^2  \end{array}
\begin{array}{c@{}} V^2 \\ NP^1  \end{array}
\\
& \rightarrow & \nonumber
\begin{array}{@{}c} N^3 \\ V^2 \end{array}
\begin{array}{@{}c} V^2 \\  D^4 \end{array}
\begin{array}{c@{}}  {} \\ N^3  \end{array}
\\
& \rightarrow & \nonumber
\begin{array}{@{}c}  N^3 \\ WASH^6 \end{array}
\begin{array}{c@{}}  MIT^5 \\ D^4  \end{array}
\begin{array}{c@{}} {} \\ N^3  \end{array}
\\
& \rightarrow & \nonumber
\begin{array}{@{}c}  PAS^7 \\ WASH^6 \end{array}
\begin{array}{c@{}}  MIT^5 \\ D^4  \end{array}
\begin{array}{c@{}} {} \\ DISH^8  \end{array}
\\
& \rightarrow & \nonumber
\begin{array}{@{}c}  PAS^7 \\ WASH^6 \end{array}
\begin{array}{c@{}}  MIT^5 \\ \text{the}  \end{array}
\begin{array}{c@{}} {} \\ DISH^8  \end{array}
\\
& \rightarrow & \nonumber
\begin{array}{@{}c}  PAS^7 \\ WASH^6 \end{array}
\begin{array}{c@{}}  \text{moy} \\ \text{the}  \end{array}
\begin{array}{c@{}} {} \\ DISH^8  \end{array}
\\
& \rightarrow & \nonumber
\begin{array}{@{}c}  PAS^7 \\ \text{Wash} \end{array}
\begin{array}{c@{}}  \text{moy} \\ \text{the}  \end{array}
\begin{array}{c@{}} {} \\ DISH^8  \end{array}
\\
& \rightarrow & \nonumber
\begin{array}{@{}c}  \text{Pasudu} \\ \text{Wash} \end{array}
\begin{array}{c@{}}  \text{moy} \\ \text{the}  \end{array}
\begin{array}{c@{}} {} \\ DISH^8  \end{array}
\\
& \rightarrow & \nonumber
\begin{array}{@{}c}  \text{Pasudu} \\ \text{Wash} \end{array}
\begin{array}{c@{}}  \text{moy} \\ \text{the}  \end{array}
\begin{array}{c@{}} {} \\ \text{dishes}  \end{array}
\end{eqnarray} 
\end{table}

The {\bf fan-out of a constituent, a nonterminal symbol, or a
precedence array} is the number of its contiguous elements.  The {\bf
fan-out of a PAV} is the sum of the fan-outs of its component
precedence arrays.  E.g., the fan-out of the PAV in
Production~\ref{depprod-eg2-alt} is $2 + 1 = 3$.  The {\bf fan-out of
a GMTG} is the maximum of the fan-outs of the PAVs in its production
rules.  A 1-GMTG with a fan-out of 1 is a CFG.

The GMTG derivation process can be represented by a derivation tree,
just like the derivation process of CFGs.  As for CFG, GMTG derivation
trees are identical to the resulting parse trees.  Several graphical
representations are common for such trees, as illustrated in
Figure~\ref{mtree}.  For example, consider a GMTG with the production
rules in Table~\ref{mtg-eg}(a).  That GMTG can derive the structure in
Figure~\ref{mtree} as shown in Table~\ref{mtg-eg}(b).  The
multidimensional perspective in Figure~\ref{mtree}(c) led us to refer
to such trees as {\bf multitrees}.

Due to the importance of lexical information in disambiguating
linguistic structure, we shall have reason to discuss lexicalized
GMTGs (LGMTGs) of the bilexical variety (L$_2$GMTGs).  In an
L$_2$GMTG, every nonterminal symbol has the form $L[t]$ for some
terminal $t \in T$ and some label $L \in \Lambda$.  $\Lambda$ is a set
of ``delexicalized'' nonterminal labels.  Intuitively, $\Lambda$
corresponds to the nonterminal set of an ordinary CFG.  The terminal
$t$ is the {\bf lexical head} of its constituent.  One nonterminal in
each component on the RHS of an L$_2$GMTG production serves as the
{\bf head-child}
of the nonterminal in the corresponding component on the LHS.  The
head-child inherits the lexical head of its parent nonterminal.  

\section{Logics for Generalized Parsing}

\subsection{Discontinuous Spans}
\label{sec:dspan}

We now introduce some notation for describing discontinuities in parse
items, and some machinery for operating on them.  Expanding on
\namecite{Johnson85}, we define a {\bf discontinuous span} (or {\bf
d-span}, for short) as a list of zero or more intervals $( {b_1},
{e_1} ; \ldots ;{b_m}, {e_m})$, where 
\bitem
\item the $b_i$ are span beginning positions and the $e_i$ are 
span ending positions, so that $b_i \leq e_i$;
\item $ e_i \leq b_{i+1}$, which means that the intervals do not overlap;
\item a d-span is {\bf proper} if all the above inequalities are
strict; i.e., each span has non-zero width and there is a gap between
each pair of consecutive intervals;
\item an empty d-span is denoted by $()$.
\eitem
As in ordinary spans, d-span boundaries range over positions between
and around the words in a text.  
Parse items have one d-span per dimension.  We shall denote vectors of
\mbox{d-spans} by $\nu, \sigma$ and $\tau$.  A d-span that pertains to
only one particular dimension $d$ is denoted with a subscript, as in
$\sigma_d$.  When a label or a d-span variable has both a superscript
and a subscript, it refers to a range of dimensions.  E.g.,
$\sigma^i_j$ is a vector of \mbox{d-spans}, one for each dimension
from $i$ to $j$.

We define two operators over \mbox{d-spans}.
\bitem
\item $+$ is the ordered concatenation operator.  Given two
\mbox{d-spans}, it outputs the union of their intervals. E.g., $(1,3;
8,9 ) + (7,8 ) = (1,3; 7,9)$. Ordered concatenation is commutative:
$\sigma + \tau = \tau + \sigma$.
\item $\wr$ is the relativization operator\footnote{\namecite{Melamed03} used
the $\otimes$ symbol for this operator, but we rename it here to avoid
confusion with this symbol's traditional use in describing
semirings.}. Given a sequence of \mbox{d-spans}, it computes the precedence
array that describes the contiguity and relative positions of their
intervals.  E.g., $ (1,3; 8,9 ) \; \wr \; (7,8 ) = [1 ; 2, 1]$,
because if these two \mbox{d-spans} were concatenated, then the result would
consist of the 1st interval of the 1st d-span, followed by a gap,
followed by the 1st interval of the 2nd d-span, followed by the 2nd
interval of the 1st d-span.  Relativization is not commutative.
\eitem
The inputs of $+$ and $\wr$ must have no overlapping intervals, or
else the output is undefined.  Both operators apply componentwise to
vectors of \mbox{d-spans}.

\subsection{Logic~C}

\label{sec:logicc}

\begin{table*}[tb]
\tcaption{Logic~C:
$D$ is the dimensionality of the grammar and $d$ ranges over
dimensions; $n_d$ is the length of the input in dimension $d$; $i_d$
ranges over word positions in dimension $d, 1 \leq i_d \leq n_d$;
$w_{d,i_d}$ are input words; $X, Y$ and $Z$ are nonterminal symbols;
$t$ is a terminal symbol; $\pi$ is a PAV; $\sigma$ and $\tau$ are
d-spans.
\label{logic-c} }
\begin{centering}
\begin{tabular}{|r|c|} \hline
\multicolumn{1}{|l|}{{\bf Term Types}} & \\
terminal items
&
$\la d, i,  t \ra$
\\
& \\
nonterminal items
&      $\left[ X^1_D; \sigma^1_D \right]$ \\ 
& \\ 
terminating productions &
\(
\begin{array}{c} \emptyset^1_{d-1} \\ X \\ \emptyset^{d+1}_D \end{array} 
\der
\begin{array}{c} \emptyset^1_{d-1} \\ t \\ \emptyset^{d+1}_D \end{array}
\)
for $1 \leq d \leq D$
\\
& \\
nonterminating productions & 
\(
X^1_D \der \Join [ \pi^1_D ]( Y^1_D \; Z^1_D )
\)
\\
& \\
\hline
\multicolumn{1}{|l|}{{\bf Axioms}} & \\
input words &
$\la d, i_d,  w_{d,i_d} \ra$ 
for $1 \leq d \leq D ,  1 \leq i_d \leq n_d$
\\
& \\
grammar terms & as given by the grammar 
\\
\hline
\multicolumn{1}{|l|}{{\bf Inference Rule Types}}  & \\
& \\
{\em Scan} component d, $1 \leq d \leq D$ &
\(
\frac{
\la d, i, t \ra
\mbox{\Large \ ,\ }
\begin{array}{@{}c} \emptyset^1_{d-1} \\ X \\ \emptyset^{d+1}_D \end{array} 
\der
\begin{array}{c@{}} \emptyset^1_{d-1} \\ t \\ \emptyset^{d+1}_D \end{array}
}{
\left[
\ba{c}
\emptyset^1_{d-1} \\
X \\
\emptyset^{d+1}_D \\
\ea
; 
\ba{c}
()^1_{d-1} \\
(i-1, i) \\
()^{d+1}_D \\
\ea
\right]
}
\)
\\
& \\
{\em Compose} &
\Large
\(
\nonumber
\frac{
\left[ Y^1_D ; \tau^1_D \right]
{\Large \ ,\ }
\left[ Z^1_D ; \sigma^1_D \right]
{\Large \ ,\ }
X^1_D \der \Join [ \tau^1_D \; \wr \; \sigma^1_D ]( Y^1_D \; Z^1_D )
}{
\left[ X^1_D ; \tau^1_D + \sigma^1_D \right]
}
\) \\
\ignore{
& \\ \hline
& \\
\multicolumn{1}{|l|}{\bf Goal}
&    $\left[ S^1_D; (0, n_d)^1_D \right]$ \\
}
& \\ \hline
\end{tabular}
\end{centering}
\end{table*}
Table~\ref{logic-c} contains Logic~C, which is a generalization of
Logic~D1C to arbitrary GMTGs in GCNF.  Parser~C is any parser based
on Logic~C.  The input to Parser~C is a tuple of $D$ parallel texts,
with lengths $n_1, \ldots, n_D$.  

The term types used by Logic~C are direct generalizations of the term
types used by Logic~D1C.  The grammar terms represent terminating and
nonterminating production rules of a GMTG in GCNF, rather than a CFG
in CNF.  The terminal items of Logic~C have the same variables as the
terminal items of Logic~D1C, plus an additional variable $d$ to
indicate the input component to which an item pertains.  Logic~C's
nonterminal items consist of a \mbox{$D$-dimensional} label vector
$X^1_D$ and a \mbox{$D$-dimensional} d-span vector $\sigma^1_D$.  The
items need \mbox{d-spans}, rather than ordinary spans, because
Parser~C needs to know all the boundaries of each item, not just the
outermost boundaries.  Since GMTGs can generate multitexts with components
of unequal length, a d-span in one component of an item might cover
more words than a d-span in another component.  In particular, some
(but not all) dimensions of a nonterminal item can be inactive, having
an empty d-span and no label. Such lower-dimensional items are
necessary for representing multitree branches that are inactive in
some components.  A typical goal item used with Logic~C would be a
constituent covering the input multitext and labeled with the
grammar's start link.
An example of such a constituent is the outermost rectangle in
Figure~\ref{mtree}(c).

Parser~C begins by firing {\em Scan} inferences, just like Parser~D1C,
but it can {\em Scan} from each of the $D$ input components.  A {\em
Scan} inference can fire for the $i$th word $w_{d,i}$ in component $d$
if that word appears in the $d$th component of the RHS of a
terminating production in the grammar.  {\em Scan} consequents have
empty spans and no labels except in the active component.

The parser can also {\em Compose} pairs of items into larger items.
The antecedents of a {\em Compose} inference can have the same number
of active components or a different number.  If both antecedents have
inactive components, then their active components may or may not be
the same.  For example, to derive the parse tree in
Figure~\ref{mtree}, Logic~C must make two inferences involving
antecedents that have no active components in common.  These are the
inferences that compose two preterminals each\footnote{The preterminal
nodes of a parse tree inferred under a GMTG in GCNF are always active
in only one component.}. If the active components of one antecedent
are a subset of the active components of the other, then the inference
asserts that some of the yield of the higher-dimensional antecedent
vanishes in translation.  An example of such an inference is the
composition that would infer the $NP/NP$ node in Figure~\ref{mtree}.

Logic~C's conditions for item composition are the ID and LP
constraints described in Section~\ref{sec:inf} generalized to possibly
discontinuous items of arbitrary dimensionality.  Both constraints now
apply componentwise to every component of the antecedents.  The LP
constraint is now expressed using the d-span relativization operator
defined in Section~\ref{sec:dspan}: Parser~C can compose two items if
the contiguity and relative order of their span intervals is
consistent with the PAV of the antecedent production rule.  Under our
new notation for production rules, the LP constraint is completely
independent of the nonterminal labels.  Such independence of
constraints is desirable for modular implementation, as well as for
concise logic specification. A complete specification of Logic~C using
the original notation for production rules would require $O(4^D)$
different {\em Compose} inference rule signatures.

Logic~C is simpler and more general than the parsing logics used by
\namecite{Wu97} and \namecite{Melamed03}.  Both the {\em Link} inference
rule in \namecite{Melamed03}'s Parser~R2D2A and Equation~1 in
\cite{Wu97} compose terminal items, but neither logic permits 
monolingual nonterminal items to compose with each other.  In
contrast, Logic~C never composes terminals, so it involves only two
types of inference rules.  However, its {\em Compose} inference rule
is more general because it admits composition of two lower-dimensional
items that are active in the same dimension, composition of two items
that are active in different dimensions, and compositions of two items
that are active in a different number of dimensions, in addition to
the usual compositions of items that are active in all dimensions.
Simplicity of description does not preclude computational complexity.
However, conceptual complexity correlates with difficulty of
engineering.  To our knowledge, there have been no studies of the
relative benefits of the two kinds of bottom-up logic.  In the absence
of evidence in support of more complicated logic, Occam's razor
supports Logic~C.

\subsection{Worst-Case Computational Complexity}
\label{sec:comcom}

The abstract parsing algorithm in Table~\ref{napa} has several sources
of computational complexity. If the simplest possible search strategy
is used (such as CKY), then the dominant source of complexity is the
logic.  We shall analyze the space and time complexity of any parser
based on Logic~C, using an extension of the static analysis method of
\namecite{McAllester02}.

The worst-case space complexity of a parser is within a constant
factor of the maximum number of possible distinct term instances that
it needs to keep track of.  A term's signature uniquely determines how
the term can combine with other terms, so two terms that have the same
values for the variables in the signature will never differ on whether
they can participate in an inference rule.  Therefore, we never need
to store more than one term with the same variable values.  The number
of unique combinations of variable values is the product of the sizes
of the variables' ranges.

For a given GMTG $G$, let $f$ be the fan-out of $G$, and let $|N|$ be
the number of nonterminal symbols in $G$.  Let $n$ be the length of
the longest component of the input multitext.  We assume that $n$ is
always smaller than the size of $G$'s terminal set.  Then the number
of possible distinct terminal items in Parser~C will be negligible
compared with the number of possible distinct nonterminal items.  The
free variables in a nonterminal item's signature are its nonterminal
symbol and span boundary in each dimension.  The maximum number of
required boundaries is exactly $2f$, and each of the boundaries can
range over $O(n)$ possible positions.  Thus, the space complexity of
Parser~C for a given $D$-GMTG $G$ is in $O(|N|^Dn^{2f})$.  If $G$ is
bilexical, then the number of possible nonterminals hides a factor of
$n^D$, raising the space complexity of Parser~C to
$O(|\Lambda|^Dn^{D+2f})$.

If the search strategy imposes an ordering of inferences that
guarantees correctness and avoids duplication of effort,\footnote{In
general, agenda-based search strategies offer no such guarantee.} then
the worst-case running time of the abstract parsing algorithm is a
product of three factors: the number of possible unique inference rule
instantiations, the computational effort required for each
instantiation\footnote{The analysis of \namecite{Melamed03} omitted
this factor.}, and an implementation-specific constant. The number of
possible unique inference rule instantiations is the product of the
sizes of the ranges of the free variables that appear in the inference
rules.  For Parser~C, these variables are the nonterminals and the
\mbox{d-spans}.  The PAVs are not free variables because they are
uniquely determined by the \mbox{d-spans}.  Assuming a fixed maximum
fan-out $f$ for the given grammar, the number of different spans in each
inference depends on how many boundaries are shared between the
antecedent items.  In the best case, all the boundaries are shared
except the two outermost boundaries in each dimension, and the
consequent is contiguous.  In the worst case, no boundaries are
shared, and the inferred item stores all the spans of the antecedent
items.  In any case, if $y$ and $z$ are the fan-out of the composed
items, and $x$ is the fan-out of the inferred item, then the number of
free boundaries in a Compose inference is $x + y + z$. Thus, in the
worst case, the number of free boundaries involved in a Compose
inference is $3f$.  Each of these boundaries can range over $O(n)$
possible values.  Thus, there are $O(n^{3f})$ possible different
\mbox{d-span} values.  There are three nonterminals per dimension,
which can have $O(|N|^{3D})$ possible different values.  Finally, each
inference rule instantiation requires the computation of the PAV in
the antecedent grammar term and the computation of the \mbox{d-span}
in the consequent, each at a cost in $O(f)$.  The total time
complexity of Parser~C is in $O(f|N|^{3D}n^{3f})$.  For a binarized
L$_2$GMTG, which also needs to keep track of two lexical heads per
dimension per inference, this complexity rises to
$O(f|\Lambda|^{3D}n^{2D+3f})$.

We presented Logic~C for its descriptive simplicity (only two inference
rule types) and familiarity (from the CKY algorithm), not for its
efficiency.  Many other parsing logics are possible, and some of them
offer lower worst-case time complexity with no loss of generality
\cite{EisnerSatta99,Melamed03}.  Nevertheless, the worst-case
computational complexity of generalized parsing will always be at least as
high as that of ordinary parsing.

\subsection{Efficiency Despite Complexity}
\label{sec:effic}

For most practical applications, monolingual parsing in $O(n^3|N|^3)$
is infeasible.  If generalized parsing is even more expensive, some
would argue, then it will never be more than a theoretical curiosity.
Yet, monolingual parsers are used daily in academia and in industry,
because the average run times of well-engineered parsers are typically
just a tiny fraction of the theoretical worst case.  The same is true
for WFST-based SMT, which involves inference algorithms with
exponential computational complexity \cite{Knight99}, and which is
nevertheless the dominant approach in the field.  Evidence is
beginning to emerge that, as for these other classes of theoretically
expensive algorithms, worst-case computational complexity should not
prevent anyone from using generalized parsers
\cite{Chiang05,DingPalmer05}.

One of the advantages of machine translation by generalized parsing is that its
practitioners need only generalize the efficiency mechanisms that have
already been developed for ordinary parsers.  The two main techniques
used to speed up parsers are pruning (also known as
``thresholding'') and outside cost estimation
\cite[e.g.\ ]{Goodman98,CaraballoCharniak98,KleinManning03}.
The parsing logics of \namecite[Chapter~5]{Goodman98} use outside cost
estimates for making decisions about pruning.  However, it is also
possible to prune without estimating outside costs.
Let us consider how these
techniques can speed up generalized parsing.

\namecite{Goodman98} augmented his parsing logics with pruning by
adding side-conditions to the inference rules.  Side-conditions on
inference rules are always boolean tests, even if the semiring
involved is not boolean.  Side-conditions used for pruning test
whether the semiring values of certain terms in the inference rule are
larger than certain other values, such as the values of certain axiom
terms (constants), the values of other terms in the same inference rule,
and/or other values recorded in the parse state.  Candidate inferences
are discarded without firing if their side-conditions are evaluated to
be false.

Side conditions involving properties of the parse state that are
neither constants nor local to the inference rule usually render the
logic nonmonotonic, perhaps even able to remove elements from SetAntSets.  If
pruning functionality is added at a sufficiently high level of
abstraction, then nonmonotonicity need not significantly increase the
difficulty of correct implementation. The side-condition test can be
added between lines 12 and 13 of the abstract parsing algorithm in
Table~\ref{napa}.  If the side-condition is false, the algorithm
removes the antecedent set from the consequent's SetAntSet instead of
adding it there, and proceeds to line 14.  Since the abstract parsing
algorithm is independent of the dimensionality of the input or the
grammar, it can apply side-conditions from logics for generalized
parsing in exactly the same way as from logics for ordinary parsing.

The outside cost estimate of a term is an estimate of the difference
between the cost of that term and the cost of a possible descendant
goal term.  A* estimates are a well-known special subclass of outside
cost estimates used in parsers.  Outside cost estimates can be used to
guide the search strategy towards terms that are more likely than
others to be on the path to the goal.  Since search strategies are
independent of the dimensionality of the input or the grammar, the
necessary modifications to the search strategy in a generalized parser
are the same as they are for the search strategy in an ordinary
parser, so we refer the reader elsewhere for details
\cite[e.g.]{KleinManning03}. The necessary modifications to the
parsing logic can vary, depending on what additional information the
search strategy needs about the state of the parse.  For example, to
compute outside costs for his monolingual bottom-up parsing logics,
\namecite{Goodman98} augmented them with new types of ``summary''
terms, which keep track of outside costs for equivalence classes of
ordinary terms.  These new term types are then used in side-conditions
to make pruning decisions.

\section{Translation}

\label{sec:trans}

A $D$-GMTG can guide a multiparser to infer the hidden structure
of a $D$-component multitext.  Now suppose that we have a $D$-GMTG and
an input multitext with only $I$ components, where $I \leq D$.  When some of
the component texts are missing, we can ask the parser to infer a
\mbox{$D$-dimensional} multitree that includes the missing components,
which are supplied by the grammar.  The
resulting multitree will cover the $I$ {\bf input
components/dimensions} among its $D$ dimensions.  It will also express
the $D-I$ {\bf output components/dimensions}, along with their
tree structures. When a parser's input can have fewer dimensions
than the parser's grammar, we call it a {\bf translator}.

\subsection{Translator~CT}

\begin{table*}
\tcaption{Logic~CT:
$D$ is the dimensionality of the grammar, $I$ is the dimensionality of
the input, and $d$ ranges over dimensions; $n_d$ is the length of the
input in dimension $d$; $i_d$ ranges over word positions in dimension
$d, 1 \leq i_d \leq n_d$; $w_{d,i_d}$ are input words; $X, Y$ and $Z$
are nonterminal symbols; $t$ is a terminal symbol; $\pi$ is a PAV;
$\sigma$ and $\tau$ are d-spans.
\label{logic-ct}}
\begin{centering}
\begin{tabular}{|r|c|} \hline
\multicolumn{1}{|l|}{{\bf Term Types}} & \\
terminal items
&
$\la d, i,  t \ra$
\\
& \\
nonterminal items
&       $\left[ X^1_D; \sigma^1_I \right]$ \\ 
& \\ 
terminating productions &
\(
\begin{array}{c} \emptyset^1_{d-1} \\ X \\ \emptyset^{d+1}_D \end{array} 
\der
\begin{array}{c} \emptyset^1_{d-1} \\ t \\ \emptyset^{d+1}_D \end{array}
\)
for $1 \leq d \leq D$
\\
& \\
nonterminating productions & 
\(
X^1_D \der \Join [ \pi^1_D ]( Y^1_D \; Z^1_D )
\)
\\
& \\
\hline
\multicolumn{1}{|l|}{{\bf Axioms}} & \\
input words &
$\la d, i_d, w_{d,i_d} \ra$ 
for $1 \leq d \leq I ,  1 \leq i_d \leq n_d$
\\
& \\
grammar terms & as given by the grammar 
\\
\hline
\multicolumn{1}{|l|}{{\bf Inference Rule Types}}  & \\
& \\
{\em Scan} component d, $1 \leq d \leq I$ &
\(
\frac{
{\Large
\la d, i, t \ra
}
\mbox{\Large \ ,\ }
\begin{array}{@{}c} \emptyset^1_{d-1} \\ X \\ \emptyset^{d+1}_I \\ \emptyset^{I+1}_D \end{array} 
{\Large \ \der \ }
\begin{array}{c@{}} \emptyset^1_{d-1} \\ t \\ \emptyset^{d+1}_I \\ \emptyset^{I+1}_D \end{array}
}{
\left[
\ba{c}
\emptyset^1_{d-1} \\
X \\
\emptyset^{d+1}_I \\
\emptyset^{I+1}_D
\ea
; 
\ba{c}
()^1_{d-1} \\
(i-1, i) \\
()^{d+1}_I \\
\mbox{ }
\ea
\right]
}
\)
\\
& \\
{\em Load} component d, $I < d \leq D$ &
\(
\frac{
\begin{array}{@{}c} \emptyset^{1}_{I} \\ \emptyset^{I+1}_{d-1} \\ X \\ \emptyset^{d+1}_D \end{array} 
{\Large \ \der \ }
\begin{array}{c@{}} \emptyset^{1}_{I} \\ \emptyset^{I+1}_{d-1} \\ t \\ \emptyset^{d+1}_D  \end{array} 
}{
\left[
\ba{c}
\emptyset^1_I \\
\emptyset^{I+1}_{d-1} \\
X \\
\emptyset^{d+1}_D \\
\ea
; 
\ba{c}
()^1_I \\
\mbox{ } \\
\mbox{ } \\
\mbox{ } \\
\ea
\right]
}
\)
\\
& \\
{\em Compose} &
\Large
\(
\nonumber
\frac{
\left[ Y^1_D ; \tau^1_I \right]
{\Large \ ,\ }
\left[ Z^1_D ; \sigma^1_I \right]
{\Large \ ,\ }
X^1_D
{\Large \ \der \Join\ }
\left[ 
\normalsize
\begin{array}{@{}c@{}}
\tau^1_I \; \wr \; \sigma^1_I \\[1mm]
\pi^{I+1}_D
\end{array}
\Large
\right] 
\left(
Y^1_D \; Z^1_D
\right)
}{
\left[ X^1_D ; \tau^1_I + \sigma^1_I \right]
} 
\) \\
\ignore{
& \\ \hline
& \\
\multicolumn{1}{|l|}{\bf Goal}
&  $\left[ S^1_D; (0, n_d)^1_I \right]$  \\
}
& \\ \hline
\end{tabular}
\end{centering}
\end{table*}

Table~\ref{logic-ct} shows Logic~CT, which is a generalization of
Logic~C.  The items of
Logic~CT have a \mbox{$D$-dimensional} label vector, as usual.
However, their d-span vectors are only \mbox{$I$-dimensional}.  Recall
that the purpose of \mbox{d-spans} is to help the parser to enforce LP
constraints, so that the input is covered only once.  It would be
pointless to constrain the absolute positions of items on the output
dimensions, because on those dimensions there is no input to cover.
On the output dimensions, we need only constrain the relative order of
items.  Constraints on the relative order are specified by
$\pi^{I+1}_D$ in the {\em Compose} grammar term, which is the part of
the PAV that pertains to the output dimensions.  

Translator~CT is any generalized parser based on Logic~CT.  
Translator~CT scans only the input components.  Terminating
productions with active output components are {\em Load}ed: Their LHSs
are added to the chart without d-span information.  Composition
proceeds as before, except that there are no constraints on the
precedence arrays in the output dimensions --- the precedence arrays
in $\pi^{I+1}_D$ are free variables.

As in Parser~C, the first few {\em Compose} inferences fired by
Translator~CT typically link items that have no active dimensions in
common.  If one of the items exists only in the input dimension(s),
and the other only in the output dimension(s), then this inference is,
de facto, translation.  As for all inference rules, the possible
translations are determined by consulting the grammar.  Thus, in
addition to its usual function of evaluating linguistic structures,
the grammar simultaneously functions as a translation model.

In summary, Logic~CT differs from Logic~C as follows: 
\bitem
\item Items store no absolute position information (\mbox{d-spans})
for the output components.
\item For the output components, the {\em Scan} inferences are
replaced by {\em Load} inferences, which are just like {\em Scan}s
except that they are not constrained by input.
\item The {\em Compose} inference does not constrain the absolute
positions of items on the output components, although the antecedent PAV
still constrains their relative positions.
\eitem
We have constructed a translator from a multiparser merely by relaxing
some constraints on the output dimensions.  Table~\ref{logic-ct} is so
similar to Table~\ref{logic-c} because Parser~C is just Translator~CT
for the special case where $I=D$. The relationship between the two
classes of algorithms is easier to see from their declarative logics
than it would be from their procedural pseudocode or equations.  

The relationship between translation and ordinary parsing was noted a
long time ago \cite{AhoUllman69}, but here we articulate it in more
detail: Ordinary parsers are a proper subclass of multiparsers, which
are a proper subclass of translators. That Logic~C is a special case
of Logic~CT explains why we view multiparsers as a subclass of
translators.  It may be counterintuitive to think of algorithms that
produce no new words as translators, but any analysis or optimization
that is valid for translators is also valid for multiparsers.  The
subclass relationship is convenient for both theoretical investigation
and practical implementation.

Logic~CT can be used with any of the semirings listed in
Section~\ref{sec:sring}.  For example, under a boolean semiring, this
logic will succeed on an \mbox{$I$-dimensional} input if and only if
it can infer a \mbox{$D$-dimensional} multitree, whose root is the
goal item.  Such a tree would contain a $(D-I)$-dimensional
translation of the input.  Thus, under a boolean semiring,
Translator~CT can determine whether a translation of the input exists,
according to the grammar.

With a probabilistic GMTG (PGMTG) and the inside semiring,
Translator~CT can compute the total probability of all $D$-dimensional
multitrees containing the \mbox{$I$-dimensional} input.  All these
derivation trees, along with their probabilities, can be efficiently
represented as a packed parse forest, rooted at the goal item.
Unfortunately, finding the most probable output string still requires
summing probabilities over an exponential number of trees.  This
problem was shown to be NP-hard in the one-dimensional case
\cite{Simaan96}.  There is no reason to believe that it is any easier
in multiple dimensions.

\subsection{Practical Variations}
\label{sec:tx-var}
\subsubsection{Other Semirings and Search Strategies}
The Viterbi-derivation semiring would be used most frequently in
practice.  Given a $D$-PGMTG, Translator~CT can use this semiring to find
the single most probable \mbox{$D$-dimensional} multitree that covers the
\mbox{$I$-dimensional} input.  For example, suppose that 
\bitem
\item all the productions in Table~\ref{mtg-eg}(a) have probability 1.0,
except that Production~(\ref{washprod}) has probability 0.7 and
Production~(\ref{cleanprod}) has probability 0.3;
\item we employ a uniform cost search strategy, so that the translator
makes inferences in order of decreasing probability of the
consequent (ties are broken according to the lexicographic
order of the consequent labels);
\item the input is {\em Pasudu moy}.
\eitem
\begin{figure*}[p]
\footnotesize
\begin{center}
\begin{psmatrix}[rowsep=1.5, colsep=0.1]
&
&
[name=S]
\mbox{$9. \left[ \begin{array}{c} S \\ S \end{array}
              ;
              \begin{array}{c} (0,2) \\ {} \end{array} \right] $
}
\\ [-10mm]
&
&
&
[name=Sder,colsep=-1]
\mbox{
$
\begin{array}{c} S \\ S \end{array}
              \der \Join 
		\left[
	      \begin{array}{@{}c@{}} {[}1,2{]} \\ {[}2,1{]} \end{array}
		\right]
		\left(
	      \begin{array}{c} NP \\ NP \end{array}
	      \begin{array}{c} V \\ V \end{array}
       \right)
$
}
\\ [-10mm]
&
[name=V]
\mbox{ 8. $\left[  \begin{array}{c}V \\ V \end{array}
                ;
		\begin{array}{c} (1,2) \\ {} \end{array}
        \right] $
}
&
[name=NP]
\mbox{ 6. $\left[  \begin{array}{c}NP \\ NP \end{array}
                ;
		\begin{array}{c}  (0,1) \\ {} \end{array}
        \right]$ }
\\ [-10mm]
&
\begin{psmatrix}[rowsep=1.5]
& &
[name=Vder,colsep=-0.5]
$
       \begin{array}{c} V \\ V \end{array}
       \der \Join
	\left[
       \begin{array}{@{}c@{}}  {[} 1 {]} \\ {[} 2 {]} \end{array}
       \right]
	\left(
       \begin{array}{c}  MIT \\ \emptyset \end{array}
       \begin{array}{c} \emptyset \\ WASH \end{array}
 \right)
$
\\ [-10mm] 
[name=WASH]
$7. \left[  \begin{array}{c} \emptyset \\ WASH \end{array}
         ;
	 \begin{array}{c} \emptyset \\ {} \end{array} 
\right]$ 
&
[name=MIT]
$1. \left[  \begin{array}{c} MIT \\ \emptyset \end{array}
         ;
	 \begin{array}{c} (1,2) \\ {} \end{array} 
\right]$
\end{psmatrix} 
\\ [-10mm]
&
[colsep=-4.2]
\begin{psmatrix}
[name=wash]
$
           \begin{array}{c} \emptyset \\ WASH \end{array}
	   \der
           \begin{array}{c}  \emptyset \\ \text{Wash} \end{array}
$
&
[name=moy]
$
         \begin{array}{c}  MIT \\  \emptyset \end{array}
	 \der
         \begin{array}{c} \text{moy} \\ \emptyset \end{array}
$
&
[name=moy-ax]
$\la 1, 2, moy \ra $
\end{psmatrix}
&
&
[name=NPder,colsep=-1.5]
$
\begin{array}{c} NP \\ NP \end{array}
\der \Join
\left[
\begin{array}{@{}c@{}}{[}1{]} \\ {[}2,1{]} \end{array}
\right] 
\left(
\begin{array}{c} N \\ N  \end{array}
\begin{array}{c} \emptyset \\ D \end{array}
\right)
$
\\
&
[name=D,colsep=5]
$3. \left[
  \begin{array}{c} \emptyset \\ D \end{array}
  ;
  \begin{array}{c} () \\ \mbox{} \end{array}
  \right]$
&
[name=N]
$
 5. \left[ \begin{array}{c} N \\ N \end{array}
  ;
  \begin{array}{c}  (0,1) \\ {} \end{array}
  \right]
$
&
\\ [-10mm]
&
[name=the,colsep=5]
$
   \begin{array}{c} \emptyset \\ D \end{array}
   \der
   \begin{array}{c} \emptyset \\ the \end{array}
$ \\ [-10mm]
&& & 
[name=Nder, colsep=-1.5]
$
  \begin{array}{c} N \\ N \end{array}
  \der \Join
  \left[
  \begin{array}{@{}c@{}} {[} 1 {]} \\ {[}2{]} \end{array}
  \right]
\left(
  \begin{array}{c} PAS \\ \emptyset \end{array}
  \begin{array}{c} \emptyset \\ DISH \end{array}
\right)
$
\\ [-10mm]
&
[name=DISH,colsep=2]
$
4. \left [
\begin{array}{c} \emptyset \\ DISH \end{array}
;
\begin{array}{c} \emptyset \\ {} \end{array}
\right]
$
&
[name=PAS,colsep=0]
$
2. \left [
\begin{array}{c}  PAS \\ \emptyset \end{array}
;
\begin{array}{c}  (0,1) \\ \mbox{} \end{array}
\right]
$
&
\\ [-10mm]
&
[name=dishes,colsep=2]
$
   \begin{array}{c}  \emptyset \\ DISH    \end{array} 
   \der
   \begin{array}{c} \emptyset \\ \text{dishes} \end{array}
$
&
    [name=pasudu,colsep=0]
$
   \begin{array}{c} PAS \\ \emptyset \end{array} 
   \der
   \begin{array}{c}\text{Pasudu} \\ \emptyset \end{array}
$
&
[name=pas-ax]
$\la 1, 1, Pasudu \ra $
\end{psmatrix}
\psset{arrows=-, arrowsize=3pt 3}
\ncline{S}{V}
\ncline{S}{NP}
\ncline{S}{Sder}
\ncline{V}{Vder}
\ncline{V}{WASH}
\ncline{V}{MIT}
\ncline{WASH}{wash}
\ncline{MIT}{moy}
\ncline{MIT}{moy-ax}
\ncline{NP}{D}
\ncline{NP}{N}
\ncline{NP}{NPder}
\ncline{D}{the}
\ncline{N}{Nder}
\ncline{N}{DISH}
\ncline{N}{PAS}
\ncline{DISH}{dishes}
\ncline{PAS}{pasudu}
\ncline{PAS}{pas-ax}
\end{center}
\caption{Proof tree for Translator~CT's inference of the multitree in
Figure~\ref{mtree}, using the GMTG in Table~\ref{mtg-eg}(a) with a
Viterbi-derivation semiring, on input {\em Pasudu moy}.  The child
nodes of each item contain its antecedents.  The nonterminal items are
numbered to indicate the order of their inference.  
\label{parser-t-eg1}}
\end{figure*}
The inferences that Translator~CT would make under these conditions
are shown in the proof tree in Figure~\ref{parser-t-eg1}.  Each
internal node represents an item.  The children of each item are its
antecedents.  The nonterminal items are numbered to indicate the order of
inference.  For example, the consequents numbered~3 and~6 precede
the one numbered~7, because the former all have probability 1.0, but
the probability of the latter is lowered by the probability of its
antecedent production rule.  The 2nd item is inferred before the 3rd
because the label ``$PAS / \emptyset$'' of the 2nd consequent
precedes the label ``$\emptyset / D$'' of the 3rd consequent in the
lexicographic order.  Note that the information in the proof tree in
Figure~\ref{parser-t-eg1} is a superset of the information in 
in Figure~\ref{mtree}.

One of the productions in Table~\ref{mtg-eg} is absent from
Figure~\ref{parser-t-eg1}.  By replacing the usual CKY search strategy
with a more sophisticated one, the translator avoids the expense of an
inference involving Production~(\ref{cleanprod}).  The benefits of
alternative search strategies are easier to see when the grammar, the
logic, the semiring, and the termination condition are abstracted away
and held constant.

\subsubsection{Other Logics}
\label{sec:otherlog}
A naive implementation of Logic~CT would be rather inefficient in
practice.  It requires {\em Load}ing an axiom for each word in the
target vocabulary, regardless of whether a {\em Load}ed word is a
possible translation of some input word.  With a large grammar, most
{\em Load} consequents would never be {\em Compose}d with input items,
so those {\em Load} inferences would be a waste of time.  A
straightforward optimization is to check whether a target word might
be the translation of some input word before {\em Load}ing it.  To
implement this optimization, replace the {\em Load} inference rule
with the following inference rule, for $I < d \leq D$:
\begin{equation}
\label{LoadCompose}
\nonumber
\frac{
\begin{array}
{@{}c} \emptyset^{1}_{I} \\ \emptyset^{I+1}_{d-1} \\  Z \\ \emptyset^{d+1}_D 
\end{array} 
{\Large \ \der \ }
\begin{array}
{c@{}} \emptyset^{1}_{I} \\ \emptyset^{I+1}_{d-1} \\ t \\ \emptyset^{d+1}_D  
\end{array} 
{\Large \ ,\ }
\left[ Y^1_D ; \tau^1_I \right]
{\Large \ ,\ }
X^1_D
{\Large \ \der \Join\ }
\left[ 
\normalsize
\begin{array}{@{}c@{}}
\tau^1_I \; \wr \; \sigma^1_I \\[1mm]
\pi^{I+1}_D
\end{array}
\Large
\right] 
\left(
Y^1_D \; 
\begin{array}
{@{}c} \emptyset^{1}_{I} \\ \emptyset^{I+1}_{d-1} \\ Z \\ \emptyset^{d+1}_D 
\end{array} 
\right)
}{
\left[ X^1_D ; \tau^1_I + \sigma^1_I \right]
}
\end{equation}
This inference rule is essentially a macro of two rules from Logic~CT:
a {\em Load} inference, and a {\em Compose} inference that could
follow the {\em Load} when a suitable antecedent item $[Y^1_D ;
\tau^1_I ]$ has been inferred from the input.  The macro will fire
once for every (input item, target word) pair, where the target word
is a possible translation of the input item, according to the grammar.
This macro admits a greater variety of inferences than the {\em Link}
inference rule of \namecite{Melamed03}, because the antecedent input
item $[Y^1_D ; \tau^1_I ]$ need not be a {\em Scan} consequent.  It
can represent an arbitrarily deep multitree over the input components.
On the other hand, this macro does not allow {\em Load} consequents to
{\em Compose} with each other before composing with items covering
some part of the input.  More sophisticated translation logics are
necessary to achieve complete flexibility efficiently.

\subsubsection{Other Grammars}
An SMT system can benefit from mixing the predictions of its
translation model with those of a more reliable monolingual language
model \cite{Brown+93}.  The classic way to mix a translation model
with a language model is the so-called noisy-channel framework.  This
framework applies to conditional models, but \namecite{Melamed04b}
shows that monolingual language models can also be mixed in a
principled way with joint models such as probabilistic synchronous
grammars.  The key benefit of such a mixture is that it can help to
evaluate every inference fired by a parsing logic.  In this manner,
the language model can greatly accelerate the translation process, in
comparison to algorithms that apply a language model only after a
complete multiforest has been inferred.  For example, the decoder of
\namecite{YamadaKnight02} first builds a forest of multitrees, and
then searches for the single most probable output in the forest using
a language model.  If only a single translation is desired, then there
is no need to compute a parse forest.  Moreover, if only the single
most probable translation is desired, then various pruning methods can
be used to speed up the search.  A PGMTG mixed with a target language
model can provide sharper term probability estimates, making the
pruning methods more efficient.

\subsection{Discussion}
The multitree inferred by the translator will have the words of both
the input and the output components in its leaves. In practice, we
usually want the output as a string tuple, rather than as a multitree.
Under the various derivation semirings \cite{Goodman98},
Translator~CT can store the output precedence arrays $\pi^{I+1}_D$ in
each internal node of the tree.  The intended ordering of the
terminals in each output dimension can be assembled from these
arrays by a linear-time {\bf linearization} post-process that
traverses the finished multitree in postorder.

To the best of our knowledge, Translator~CT is the first to be
compatible with all of the semirings listed in
Section~\ref{sec:sring}, among others.  It is also unique in being
able to accommodate multiple input components and multiple output
components simultaneously.  When a source document is available in
multiple languages, a translator can benefit from the disambiguating
information in each.  Translator~CT can take advantage of such
information without making the strong independence assumptions of
\namecite{OchNey01}.  When Translator~CT is used to translate into
multiple languages simultaneously, each translation is constrained not
only by the input, but also by all the other translations.  This
approach might effect greater consistency across output components,
which is one of the putative benefits of the interlingual approach to
MT. Indeed, the language\footnote{Here we intend the formal language
theory sense of ``language.''} of multitrees can be viewed as an
interlingua.

\section{Hierarchical Alignment}
\label{sec:align}

In Section~\ref{sec:trans} we explored inference of
\mbox{$I$-dimensional} multitrees under a \mbox{$D$-dimensional}
grammar, where $D \geq I$.  Now we generalize along the other axis of
Figure~\ref{parsematrix}(a).  It is often useful to infer
\mbox{$I$-dimensional} multitrees without the benefit of an
\mbox{$I$-dimensional} grammar.  One application is inducing a parser in one
language from a parser in another \cite{Lu+02}.  The application that
is most relevant to this article is bootstrapping an \mbox{$I$-dimensional}
grammar.  

In theory, it is possible to estimate a PGMTG from multitext in an
unsupervised manner,
starting with a random or uniform distribution over production rules.  However, the 
quality of the parameter estimates 
is greatly affected by how they are initialized, so such a simple
approach is unlikely to produce good results.  A more reliable way to
estimate PGMTG production probabilities is from a corpus of multitrees
--- a {\bf multitreebank}.\footnote{In contrast, a parallel treebank
\cite{Han+01} might contain no information about translational
equivalence beyond sentence alignments.}  Despite some recent efforts
to manually construct multitreebanks \cite{Uchimoto+04}, it is unlikely
that they will become available for more than a handful of language
pairs any time soon.
The most straightforward way to create a multitreebank is to parse
some multitext using a multiparser, such as Parser~C.  However, if the
goal is to bootstrap an $I$-PGMTG, then there is no $I$-PGMTG that can
evaluate the grammar terms in the parser's logic.

Our solution is to orchestrate lower-dimensional knowledge sources to
evaluate the grammar terms.  Then, we can use our favorite
multiparsing logic to align multitext into a multitreebank.
If we have no PGMTG, then we can use other criteria to evaluate
inferences.  These other criteria can be based on various subsets of
the information available in inference rules.

For example, given a tokenized set of tuples of parallel sentences, it
is always possible to estimate a word-to-word translation model
$\Pr(u^{D+1}_I | u^1_D)$\cite{Brown+93}\footnote{Although
most of the literature discusses word translation models between only
two languages, it is possible to use one language as a pivot to
combine several 2D models into a higher-dimensional model
\cite{MannYarowsky01}.}.  Such a probability distribution ranges over
parts of the nodes of multitrees.  Even if we have no basis for
choosing among different tree structures, we can prefer
multitrees whose individual nodes have higher probability.
\namecite{Chiang05} generalized this idea to bootstrap a synchronous
grammar from a pre-existing phrase-to-phrase translation model.

Research on hierarchical alignment has a rich history in the context
of example-based machine translation.  To our knowledge, all the
algorithms presented in that context presume that parse trees are
available for all multitext components, which is why that subclass of
alignment algorithms is usually called {\bf tree alignment}
\cite{Meyers+96} or {\bf structural matching} \cite{Matsumoto+93}.
The idea that alignment can be carried out under much more varied
conditions was first put forth by \namecite{Wu95b}, and further
expounded by \namecite{Wu00}.  In this section, we offer a more
precise characterization of the relationship between multiparsing and
hierarchical alignment, by showing that hierarchical alignment can be
carried out using {\em exactly} the same logics, semirings, search
strategies, and termination conditions as ordinary multiparsing
algorithms.  A generalization of what counts as a grammar is
sufficient.

\subsection{A Common Scenario}

For an extended example, we consider the common alignment scenario
where a lexicalized monolingual grammar is available for just one
component.  For example, many multitexts have at least one component
in one of the languages for which treebanks have been
built\footnote{At the time of writing, we are aware of treebanks for
English, Spanish, French, German, Chinese, Czech, Arabic, and
Korean.}.  Given a treebank in the language of one of the input
components, we can induce a lexicalized PCFG.
Alternatively, if a non-probabilistic parser is available for one of
the input components, then we can first parse that component, and then
proceed as we would from a treebank.  Regardless of how we obtain it,
a monolingual lexicalized grammar can guide our search for the
multitree with the most probable structure in the
resource-rich component.  More generally,\footnote{Recall that PCFGs
are a subclass of PGMTGs.} we might have a lexicalized PGMTG in $D$
dimensions, from which we want to align \mbox{$I$-dimensional}
multitrees, $I \geq D$.  Without loss of generality, we shall let the
PGMTG range over the first $D$ components.  We shall then refer to the
$D$ structured components and the $I-D$ unstructured components.

\begin{figure}
\centerline{\psfig{figure=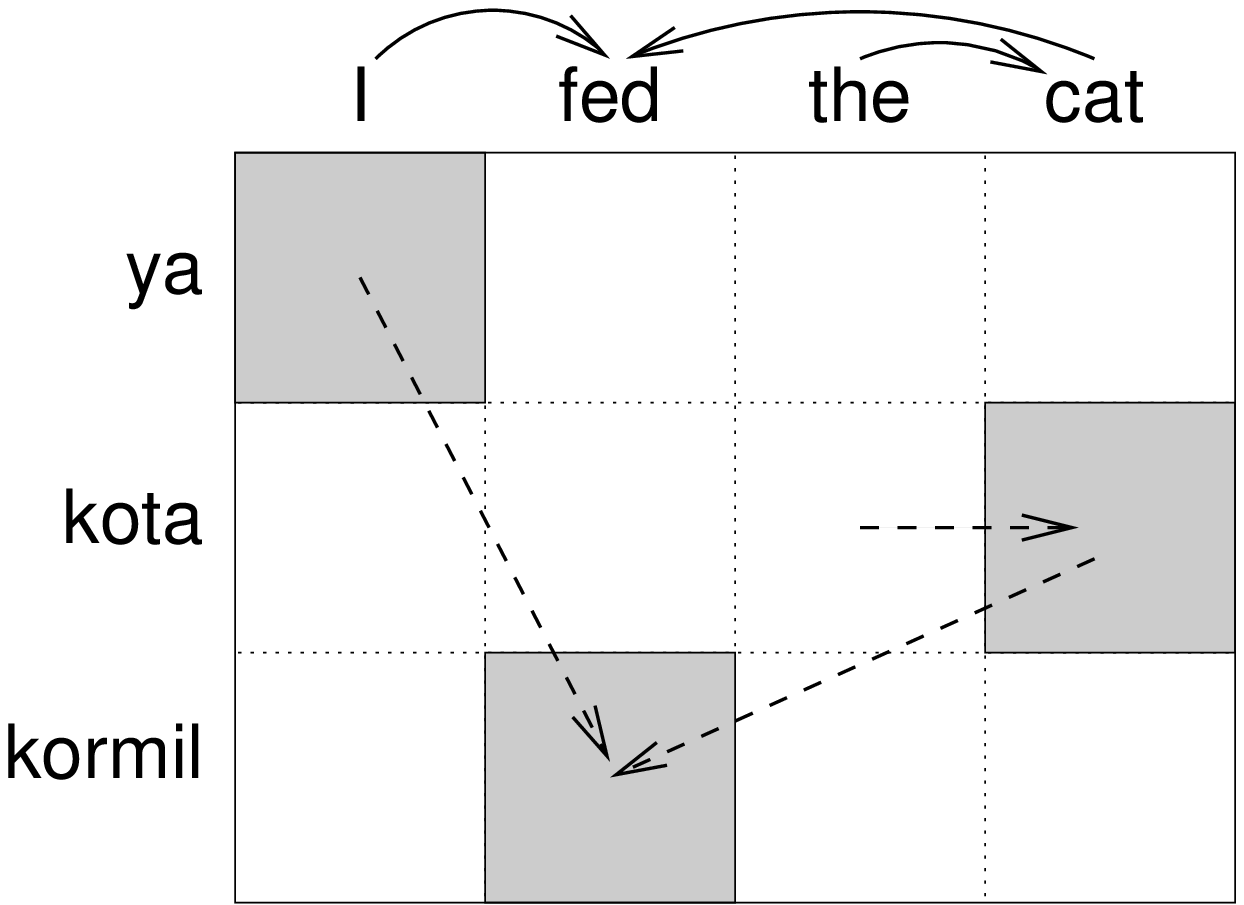,height=1in}}
\caption{
Example of constraints in hierarchical alignment. Only one synchronous
dependency structure (dashed arcs) is compatible with the monolingual
structure (solid arcs) and word alignment (shaded squares).
\label{synch-eg}} 
\end{figure}

Given a one-to-one matching between the words in a multitext, choosing
the optimal structure for one component is tantamount to choosing the
optimal synchronous structure for all components.  For example, in
Figure~\ref{synch-eg}, a monolingual grammar has allowed only one
synchronous dependency structure on the English side, and a word-to-word
translation model has allowed only one word alignment.  Ignoring the
nonterminal labels, only one dependency structure is compatible with
these constraints.  Unmatched nodes in the structured component can be
ignored.  Unmatched nodes in the unstructured component can be
heuristically attached either to the left or to the right
\cite{Wu95b}, or even randomly.  More generally, the given word
matching need not be one-to-one and the structure given for the
structured component need not be a single tree or a tree at all.
Missing substructures and other ambiguities in these input constraints
can be resolved during the alignment process.

To combine structural and translational constraints for alignment in
this manner, it is convenient to suppose that we are inducing a {\em
bilexical} PGMTG under the Viterbi-derivation semiring. Given a
bilexical PCFG, or a functionally equivalent approximation thereof, we
can search for a multitree that simultaneously has a high-probability
tree structure and a high-probability correspondence among words in
its nodes.  Such an inference process is, by definition, a generalized
parser.  It can be based on any parsing logic, including Logic~C.  If
we have no $I$-PGMTG, then we can evaluate the grammar terms in a way
that does not rely on it.
Let $G()$ be the function that the grammar uses to assign
probabilities to production rules.  Ordinarily, we have $G(LHS \der
RHS) = \Pr(RHS | LHS)$.  A modified definition is necessary in the
typical alignment scenario where the grammar has no estimates
for $\Pr(RHS | LHS)$.  

We begin with terminating productions.  For the
structured components, we retain the usual definition. I.e.,
$G(X_d[h_d] \der h_d) = \Pr(h_d | X_d[h_d])$, where the latter
probability can be looked up in a pre-existing $D$-PGMTG.  For the unstructured
components, there are no useful nonterminal labels.  Therefore, we
assume that the unstructured components use only one (dummy)
nonterminal label $\lambda$, so that $G(X_d[h_d] \der h_d) = 1$ if $X
= \lambda$ and 0 otherwise.

Our treatment of nonterminating productions follows the standard approach
of applying the chain rule for conditional probabilities and then
making independence assumptions until all the terms are in a form that
can be estimated from data.  Readers who are not interested in the
details can skip ahead to Equation~\ref{G_N_final}.  According to the
chain rule,\footnote{The procedure is analogous when the head-child is the
first nonterminal link on the RHS, rather than the second.
Information about which nonterminal link is the head-child can be
encoded in the nonterminal labels.}
\begin{eqnarray}
G(X^1_I[h^1_I] \der \Join [ \pi^1_I] ( Y^1_I[g^1_I] \; Z^1_I[h^1_I]) )
& = & \Pr(\pi^1_I, g^1_I, Y^1_I, Z^1_I | X^1_I, h^1_I) \\
& = & \Pr(\pi^1_D, g^1_D, Y^1_D, Z^1_D | X^1_I, h^1_I) \\
& \times &
\Pr(Y^{D+1}_I, Z^{D+1}_I | \pi^1_D, g^1_D, Y^1_D, Z^1_D, X^1_I, h^1_I) \\
& \times &
\Pr(g^{D+1}_I | \pi^1_D, g^1_D, Y^1_I, Z^1_I, X^1_I, h^1_I) \\
& \times &
\Pr(\pi^{D+1}_I | \pi^1_D, g^1_I, Y^1_I, Z^1_I, X^1_I, h^1_I)
\end{eqnarray}
Our first independence assumption is that the structured components of
the production's RHS are conditionally independent of the unstructured
components of its LHS:
\begin{equation}
\label{as1}
\Pr(\pi^1_D, g^1_D, Y^1_D, Z^1_D | X^1_I, h^1_I)
= \Pr(\pi^1_D, g^1_D, Y^1_D, Z^1_D | X^1_D, h^1_D) 
\end{equation}
The above probability can be looked up in the pre-existing $D$-PGMTG.
Second, since we have no useful nonterminals in the unstructured
components, we let
\begin{equation}
\label{as2}
\Pr(Y^{D+1}_I, Z^{D+1}_I | \pi^1_D, g^1_D, Y^1_D, Z^1_D, X^1_I, h^1_I) = 
\left\{ 
\begin{array}{l}
1 \text{ if } Y^{D+1}_I = Z^{D+1}_I = \lambda^{D+1}_I
\\
0 \text{ otherwise }
\end{array}
\right. .
\end{equation}
Third, we assume that the word-to-word translation probabilities are
independent of anything else:
\begin{equation}
\label{as3}
\Pr(g^{D+1}_I | \pi^1_D, g^1_D, Y^1_I, Z^1_I, X^1_I, h^1_I) = 
\Pr(g^{D+1}_I | g^1_D) \\
\end{equation}
In a typical alignment scenario, these probabilities would be
obtained from a word-to-word translation model, which would be
estimated under such an independence assumption.
Finally, we assume that the output
precedence arrays are independent of each other and uniformly
distributed, up to some maximum fan-out $f$.  Let $\mu(f)$ be
the number of unique precedence arrays of fan-out $f$ or less.
Then
\begin{equation}
\label{as4}
\Pr(\pi^{D+1}_I, | \pi^1_D, g^1_I, Y^1_I, Z^1_I, X^1_I, h^1_I) =
\Pr(\pi^{D+1}_I) = 
\prod_{d=D+1}^I \frac{1}{\mu(f)} =  \frac{1}{\mu(f)^{I-D}}.
\end{equation}
Under Assumptions~\ref{as1}--\ref{as4},
\begin{equation}
\label{G_N_final}
G(X^1_I[h^1_I] \der \Join [ \pi^1_I ] ( Y^1_I[g^1_I] \; Z^1_I[h^1_I])) = 
\frac{\Pr(\pi^1_D, g^1_D, Y^1_D, Z^1_D | X^1_D, h^1_D)
\cdot \Pr(g^{D+1}_I | g^1_D)}{\mu(f)^{I-D}}
\end{equation}
if $Y^{D+1}_I = Z^{D+1}_I = \lambda^{D+1}_I$ and 0 otherwise.  The
first term in the numerator comes from a $D$-GMTG, and the second term
from a conditional word-to-word translation model.  The denominator is
a normalization constant.

In the most common case that the multitext is just a bitext, and we
have a structured language model for just one of its components, the above
equation boils down to
\begin{equation}
\label{G_N_final_2D}
G(X^1_2[h^1_2] \der \Join [ \pi^1_2 ] ( Y^1_2[g^1_2] \; Z^1_2[h^1_2])) = 
\frac{\Pr(\pi_1, g_1, Y_1, Z_1 | X_1, h_1) \cdot \Pr(g_2 | g_1)}{\mu(f)}
\end{equation}
if $Y_2 = Z_2 = \lambda$ and 0 otherwise. We can use these estimates in
the inference rules of Logic~C, under any probabilistic semiring.

More sophisticated methods of hierarchical alignment are certainly
possible.  For example, we could project a part-of-speech tagger
\cite{Yarowsky+01} to improve our estimates in Equation~\ref{as2}.  Or
we could constrain each component with its own monolingual parse tree
\cite{SmithSmith04}.  Yet, despite their
relative simplicity, the above methods for estimating production rule
probabilities use all of the available information in a consistent
manner, without double-counting.  Bootstrapping a
PGMTG from a lower-dimensional PGMTG and a word-to-word translation
model is similar in spirit to the way that regular grammars can help
to induce CFGs \cite{LariYoung90}, and the way that simple translation
models can help to bootstrap more sophisticated ones \cite{Brown+93}.

\subsection{Word Alignment}
\label{sec:wordalign}
\label{sec:w2w}

A degenerate subclass of hierarchical alignment algorithms is
algorithms that produce flat structures, where every leaf is a child
of the root.  This subclass includes some algorithms for word
alignment. A translation lexicon (weighted or not) can be viewed as a
degenerate GMTG (not in GCNF) where every production has the form
\begin{equation}
\begin{array}{@{}c@{}} S \\ \vdots \\  S \end{array}
\der 
\begin{array}{@{}c} t_1 \\ \vdots \\ t_D \end{array}
\end{equation}
I.e., each production rewrites the start link into one terminal per
component.  Under such a GMTG, the logic of word alignment is the one
in \namecite{Melamed03}'s Parser~A.   However, instead of a
single goal item, the goal of word alignment is any {\em set} of items
that covers the input exactly once.  Also, since
nonterminals do not appear on the RHS of production rules, {\em
Compose} inferences are impossible and unnecessary, so they can be
removed from the logic if desired.  

\section{Parameter Estimation}
\label{sec:train}
\label{sec:paramest}

As for other probabilistic grammar formalisms, different parameter
estimation methods are possible for PGMTGs.  The traditional method
for PCFGs is the Inside-Outside algorithm \cite{Baker79}, which
performs unsupervised maximum likelihood estimation.  Here we present
a generalization of the logic behind this algorithm to PGMTGs in GCNF.
Our generalization can also be used to compute some common
approximations to maximum likelihood.

The Inside-Outside algorithm iterates over two stages.  The first
stage computes inside and outside item values.  The second stage
aggregates and normalizes these values to update the grammar.
\namecite{Goodman98} introduces the terms {\bf forward value} and {\bf
reverse value} as generalizations of ``inside value'' and ``outside
value'', respectively, for arbitrary semirings.  The previous section
described computation of forward values in terms of a parsing logic,
which is a generalization of \namecite{Goodman98}'s bottom-up logic
for monolingual parsing.  For computing reverse values,
\namecite{Goodman98} offers an equation, which we re-express here in
terms of forward values~$V()$ and reverse values~$Z()$:
\begin{equation}
\label{outsideval}
Z(Y_j) = \bigoplus_{\ba{c} X, Y_1, \ldots, Y_k, 1 \leq j \leq k \\
\mbox{such that } \frac{Y_1, \ldots, Y_k}{X} \ea} Z(X) \otimes
\bigotimes_{1 \leq i \leq k, i \neq j} V(Y_i)
\end{equation}

\namecite[Section~5.1]{Goodman98} stated that ``we cannot compute the
outside probability of a nonterminal until we are finished computing
all of the inside probabilities.''  However, Equation~\ref{outsideval}
shows that, in general, it is possible to compute reverse values
before computing all forward values.  The only values that are
necessary for computing the reverse value of an item are the reverse
value of the item's parent and the forward values of the item's
siblings.  Forward values of the item's parent or descendants are not
required.  For example, it is possible to compute the reverse value of
the NP in Figure~\ref{mtree} as soon as the forward value of the V is
known, without having computed the forward value of the S, the D, or
the N.

It is possible to elaborate the abstract parsing algorithm in
Table~\ref{napa} so that it computes reverse values using
Equation~\ref{outsideval}.  E.g., it could compute them after all
forward values have been computed, as suggested by Goodman.  It could
also compute them opportunistically, as soon as it knows the reverse
value of the consequent $X$ and the forward values of all the
antecedents $Y_1, \ldots, Y_k, i \neq j$.  However, the question of
when to compute term values is a question of search strategy.  In
keeping with this article's method of analysis, we abstract away the
search strategy and specify only the computational dependencies
between item values. Parsing logics are the natural way to express
such dependencies. Table~\ref{logic-cr} shows Logic~CR, which can
compute both forward and reverse term values.  In addition to
admitting a variety of search strategies, this logic admits all the
parsing semirings studied by Goodman.  It can therefore work with the
{\em unmodified} abstract parsing algorithm in Table~\ref{napa}.

\begin{table*}
\tcaption{Logic~CR:
$D$ is the dimensionality of the grammar and $d$ ranges over
dimensions; $n_d$ is the length of the input in dimension $d$; $i_d$
ranges over word positions in dimension $d, 1 \leq i_d \leq n_d$;
$w_{d,i_d}$ are input words; $X, Y$ and $Z$ are nonterminal symbols;
$t$ is a terminal symbol; $\pi$ is a PAV; $\nu, \sigma$ and $\tau$ are
d-spans.
 \label{logic-cr}}
\begin{centering}
\begin{tabular}{|r|c|} \hline
\multicolumn{1}{|l|}{{\bf Term Types}} & \\
terminal items
&
$\la d, i,  t \ra$
and
$\la d, i,  t \ra^R$
\\
& \\
nonterminal items
&      
$\left[ X^1_D; \sigma^1_D \right]$ and $\left[ X^1_D; \sigma^1_D \right]^R$\\
& \\ 
terminating productions &
\(
\begin{array}{c} \emptyset^1_{d-1} \\ X \\ \emptyset^{d+1}_D \end{array} 
\der
\begin{array}{c} \emptyset^1_{d-1} \\ t \\ \emptyset^{d+1}_D \end{array}
\)
for $1 \leq d \leq D$
\\
& \\
nonterminating productions & 
\(
X^1_D \der \Join [ \pi^1_D ]( Y^1_D \; Z^1_D )
\)
\\
& \\
\hline
\multicolumn{1}{|l|}{{\bf Axioms}} & \\
input words &
$\la d, i_d,  w_{d,i_d} \ra$ 
for $1 \leq d \leq D ,  1 \leq i_d \leq n_d$
\\
& \\
grammar terms & as given by the grammar 
\\
& \\
pivot &
\(
\nonumber
\left[ S^1_D; (0, n_d)^1_D \right]^R
\) \\
& \\
\hline
\multicolumn{1}{|l|}{{\bf Inference Rule Types}}  & \\
{\em Scan} 
& 
\(
\frac{
\la d, i, t \ra
\mbox{\Large \ ,\ }
\begin{array}{@{}c} \emptyset^1_{d-1} \\ X \\ \emptyset^{d+1}_D \end{array} 
\der
\begin{array}{c@{}} \emptyset^1_{d-1} \\ t \\ \emptyset^{d+1}_D \end{array}
}{
\left[
\ba{c}
\emptyset^1_{d-1} \\
X \\
\emptyset^{d+1}_D \\
\ea
; 
\ba{c}
()^1_{d-1} \\
(i-1, i) \\
()^{d+1}_D \\
\ea
\right]
}
\)
\\
& \\
{\em Forward Compose} &
\Large
\(
\nonumber
\frac{
\left[ Y^1_D ; \tau^1_D \right]
{\Large \ ,\ }
\left[ Z^1_D ; \sigma^1_D \right]
{\Large \ ,\ }
X^1_D \der \Join [ \tau^1_D \; \wr \; \sigma^1_D ]( Y^1_D \; Z^1_D )
}{
\left[ X^1_D ; \tau^1_D + \sigma^1_D \right]
}
\) \\
& \\
{\em Reverse Compose Right} &
\Large
\(
\nonumber
\frac{
\left[ Y^1_D ; \tau^1_D \right]
{\Large \ ,\ }
\left[ X^1_D ; \nu^1_D \right]^R
{\Large \ ,\ }
X^1_D \der \Join [ \nu^1_D \; \oslash \; \tau^1_D ]( Y^1_D \; Z^1_D )
}{
\left[ Z^1_D ; \nu^1_D - \tau^1_D \right]^R
}
\) \\
& \\
{\em Reverse Compose Left} &
\Large
\(
\nonumber
\frac{
\left[ Z^1_D ; \sigma^1_D \right]
{\Large \ ,\ }
\left[ X^1_D ; \nu^1_D \right]^R
{\Large \ ,\ }
X^1_D \der \Join [ \nu^1_D \; \oslash \; \sigma^1_D ]( Y^1_D \; Z^1_D )
}{
\left[ Y^1_D ; \nu^1_D - \sigma^1_D \right]^R
}
\) \\
& \\
{\em Reverse Scan} 
& 
\(
\frac{
\left[
\ba{c}
\emptyset^1_{d-1} \\
X \\
\emptyset^{d+1}_D \\
\ea
; 
\ba{c}
()^1_{d-1} \\
(i-1, i) \\
()^{d+1}_D \\
\ea
\right]^R
\mbox{\Large \ ,\ }
\begin{array}{@{}c} \emptyset^1_{d-1} \\ X \\ \emptyset^{d+1}_D \end{array} 
\der
\begin{array}{c@{}} \emptyset^1_{d-1} \\ t \\ \emptyset^{d+1}_D \end{array}
}{
\la d, i, t \ra^R
}
\)
\\
\ignore{
& \\ \hline
& \\
\multicolumn{1}{|l|}{\bf Goal}
&    $\left[ S^1_D; (0, n_d)^1_D \right]$ and 
\(
\la d, i, w_{d,i} \ra^R
\) 
for $1 \leq d \leq D ,  1 \leq i_d \leq n_d$
\\
}
& \\ \hline
\end{tabular}
\end{centering}
\end{table*}

The main novelty of Logic~CR is its treatment of ``reverse'' items as
a first-class term type.  The reverse items and reverse inference
rules of Logic~CR are defined so that $Z(x) = V(x^R)$.  Thus, instead
of using Equation~\ref{outsideval}, the reverse value of an item is
computed by Equation~\ref{insideval} as the forward value of the
corresponding reverse item.
The benefit of this
treatment is that computations of reverse item values can be subject
to the same kinds of optimization that are used to speed up
computation of forward values, including pruning and logic
transformations, such as the one proposed by \namecite{Melamed03}.

Let us consider how Logic~CR extends Logic~C.  It has two new term
types for recording the reverse values of terminal and nonterminal
items.  Reverse terminal items are useful for at least two purposes.
First, if the input is nondeterministic, such as a word lattice coming
from the acoustic module of a speech recognizer, then reverse values
can be useful for pruning the lattice.  Second, it is straightforward
to generalize Logic~CR into a logic for translation, the same way that
Logic~C was generalized to Logic~CT.  Then, reverse values of
terminals in the output dimensions could be used to prune and reorder
items on the agenda, the same way that a target language model is used
in WFST-based SMT.  Interestingly, a reverse terminal item can involve
any terminal in the grammar, which may or may not correspond to any
word in the input.  It is perfectly valid to compute reverse values
for partial parses whose forward value remains at its initial default
(e.g.\ probability zero).  If such values are unnecessary for the
application at hand, then their computation can be avoided using logic
optimizations analogous to the macro inference rule in
Section~\ref{sec:otherlog}.

Logic~CR also introduces a new kind of axiom called a {\bf pivot},
which declares the reverse value of the item that spans the whole
input and has the grammar's start symbol as its label.  It is
impossible to infer this value, because computation of an item's
reverse value requires knowing its parent's reverse value, and an item
spanning the whole input cannot have a parent.  Fortunately, it is
 unnecessary to compute this value, because the reverse value of
any item labeled with the grammar's start symbol is always the
multiplicative identity of the semiring.

Logic~CR has new rules for inferring the new item types.
Two {\em Reverse Compose} rules are required: one for the case where
the consequent label comes first on the RHS of the antecedent grammar
term, and one for the case where it comes second.  The computations of
the PAV in the antecedent and the d-span in the consequent of the {\em
Reverse Compose} rules involve two new operators, which perform the
inverse operations of $+$ and $\wr$ over d-spans:
\bitem
\item $-$ is the subtraction operator:  Given an ordered pair of
\mbox{d-spans} $\nu$ and $\tau$, it outputs the d-span $\sigma$ such that
$\sigma + \tau = \nu$.  The output is undefined if $\tau$ contains
intervals not covered by $\nu$.
\item $\oslash$ is the reverse relativization operator, defined by the
equation $\nu \oslash \tau = (\nu - \tau) \wr \tau$.
\eitem
Both operators apply componentwise to vectors of \mbox{d-spans}.  

With an inside semiring, Logic~CR can compute the inside and outside
item values required for a multidimensional Inside-Outside algorithm.
This algorithm can be used to re-estimate the parameters of a PGMTG in
GCNF.  Equations for aggregation and normalization are necessary to
complete the specification of the algorithm.  
Let $V^q([\iota])$ be the value of item $[\iota]$ on iteration $q$.
Let $G^q(P)$ be the value assigned by the grammar to production rule
$P$ on iteration $q$.  Let $T$ be the RHS of a terminating production
rule.  Let $I$ be a vector of word positions and let \underline{1} be
a vector of 1's.  Then the update equations are\footnote{We omit the
dimension indexes to reduce clutter.}:
\begin{equation}
\label{newtermprod}
G^{q+1}(X \der T) =
\frac{
\sum_{I s.t. w_I = T} V^q([ X ; (I - \underline{1}, I) ]) \cdot V^q([ X ; (I - \underline{1}, I) ]^R)
}{
\sum_{I} V^q([ X ; (I - \underline{1}, I) ]) \cdot V^q([ X ; (I - \underline{1}, I) ]^R)
}
\end{equation}
\begin{eqnarray}
\label{newntprod}
\lefteqn{G^{q+1}(X \der \Join [ \tau \; \wr \; \sigma ]( Y \; Z )) = }
\nonumber \\
& & \frac{
\sum_{\tau, \sigma} V^q([ X ; \tau + \sigma ]^R) \cdot V^q([Y ; \tau])
\cdot V^q([Z ; \sigma]) \cdot
G^{q}(X \der \Join [ \tau \; \wr \; \sigma ]( Y \; Z ))
}{
\sum_{\tau, \sigma} V^q([ X ; \tau + \sigma ]) \cdot V^q([ X ; \tau + \sigma ]^R)
}
\end{eqnarray}
To aggregate over multiple training sentence tuples, augment word
positions and span boundaries to record the sentence tuple number.

Parsing under the inside semiring requires summing over all possible
derivations of the training data, which precludes the efficiency
mechanisms suggested in Section~\ref{sec:effic}.  Given the
computational expense of exhaustive multiparsing, cheaper
approximations are often desirable.  Instead of computing over all
possible derivations, we can use only the $n$ best derivations for
some fixed maximum $n$.  This approach was also suggested by
\namecite{Brown+93} for the more sophisticated of their translation
models.  Logic~CR can compute this approximation without modification
if it is used with the Viterbi semiring or it's $n$-best
generalization.  The above update equations are appropriate
regardless of which of these semirings is used to compute the values
$V()$.  It is also possible to use a variant of the above equations
when $G()$ ranges over values in an expectation semiring
\cite{Eisner02}.  Such a variant, together with Logic~CR, could
compute the expected feature counts necessary to re-estimate a maximum
entropy synchronous grammar of the kind used by \namecite{Chiang05}.

Our development of Logic~CR was motivated by parameter estimation for
PGMTGs.  \namecite[Section~2.4]{Goodman98} suggests several other
applications of reverse semiring values:
\bitem
\item pruning,
\item defining non-standard criteria for parser performance, and then
\item improving parser performance on those criteria, which can result in
\item faster parsing, even without pruning.
\eitem
In all these applications it is useful to know reverse item
values before all the forward values are known.

In addition to the usual ``root'' goal item, the termination
conditions for a typical application of Logic~CR
would involve the {\em set} of reverse terminal items that correspond
to the input words.
Loosely speaking, Logic~CR would aim to reach the root goal
bottom-up, and then return to the input that it started from
top-down.  We therefore refer to the class of logics that infer both
forward and reverse values as {\em round-trip parsing logics}.

\section{Translation Evaluation by generalized parsing}

\label{sec:eval}

In recent years, it has become {\em de rigueur} to evaluate MT systems
objectively, using automated comparison with reference translations
\cite{Thompson91,BrewThompson94}.  All the currently popular
evaluation measures compute some form of string similarity.  It is not
difficult to imagine how such measures can miscalculate.  For example,
suppose that the reference translation is (R) below, and that two MT
systems output the translations (T1) and (T2).
\be
\item[(R)] Pat asked Sandy on Friday about the man from Oslo.
\item[(T1)] On Friday, Pat asked Sandy about the man from Oslo.
\item[(T2)] Pat from Oslo asked Sandy on Friday about the man.
\ee
The sentences in this example are neither long nor complicated.  Yet
all of the currently popular automatic evaluation methods would
incorrectly assign a higher score to (T2) than to (T1), because (T2)
has a longer matching \mbox{$n$-gram} with~(R).  The problem is that
string similarity is only a crude approximation to conceptual
similarity.  Methods that measure the grammaticality of translations
independently of a reference translation (e.g.,
\cite{RajmanHartley01}) are also incapable of making the desired
distinctions --- (T1) and (T2) are equally grammatical.

More sophisticated MT systems will require more sophisticated
evaluation methods.  In order to correctly evaluate examples like the
one above, an evaluation method needs a catalogue of the syntactic
alternations that preserve the meaning of an utterance.  Synchronous
grammars offer a perspicuous way to describe such alternations.  For
example, the production
\begin{equation}
\ba{@{}c@{}} \text{NP} \\ \text{NP} \ea 
\der \Join
\left[
\ba{@{}c@{}} {[1,2,3]} \\ {[1,3,2]} \ea
\right]
\left(
\ba{@{}c@{}} \text{NN } \text{PP}_1 \text{ PP}_2 \\  \text{NN }
\text{PP}_1 \text{ PP}_2 \ea
\right)
\end{equation}
could be included, to allow prepositional phrases modifying the same
head to switch places.  However, the relative order of determiners and
adjectives in English noun phrases is strict, so the production
\begin{equation}
\ba{@{}c@{}} \text{NP} \\ \text{NP} \ea 
\der \Join
\left[
\ba{@{}c@{}} {[1,2,3]} \\ {[2,1,3]} \ea
\right]
\left(
\ba{@{}c@{}} \text{Det Adj NN} \\  \text{Det Adj NN} \ea
\right)
\end{equation}
would not be included.  The grammar would also include productions
that have identical components in both dimensions.

Given such a grammar $G$, a reference translation $R$, and an MT system
output $T$, a multiparser can attempt to find a multitree
covering the bitext $(R, T)$ under $G$.  If the parser succeeds, then,
according to the grammar, $T$ is a valid translation (actually,
paraphrase) of $R$.  If the parser fails, then $T$ is not
an acceptable paraphrase of $R$, either because it does not mean the
same thing or because it is ungrammatical.

There are two practical problems with this approach.  First, it is
usually desirable to obtain a numerical grade of translation quality,
rather than just a boolean indicator of acceptability.  Second, it
would probably be infeasible, or at least unreliable, to compile all
the valid syntactic alternations manually.  One possible solution to
these problem was proposed by \namecite{Leusch+03}, who restricted
themselves to a Bracketing Transduction Grammar \cite{Wu95b} with just
one dummy nonterminal, partitioned its possible production rules into
seven classes, and manually assigned a cost for each class.

An alternative approach is to estimate the required grammar
empirically. The Linguistic Data Consortium has recently published
several ``multiple-translations'' corpora.  These are corpora
containing multiple independent translations of a set of source
documents, aligned at the sentence level.  Each set of independent
translations can be viewed as mutual paraphrases \cite{Pang+03}.  We
can estimate a monolingual PGMTG\footnote{I.e., a PGMTG that generates
the same language in all components.} from these sets of parallel sentences
using exactly the same algorithms that we use to estimate multilingual
PGMTG, as described in Sections~\ref{sec:align}
and~\ref{sec:paramest}.  Using such a PGMTG, a {\em probabilistic}
multiparser can return the {\em probability} that a translation is
valid with respect to a reference.  Different translations and the MT
systems that output them can be compared on these scores.

MT evaluation by means of a monolingual PGMTG has two advantages over
string-based methods \cite{Melamed95,Papineni+02,Melamed+03}.  First,
this method can be sensitive to meaning-preserving syntactic
alternations.  To the extent that human judges use such information in
evaluating MT outputs, an automatic evaluation method that uses such
information might do a better job of predicting human judgments.
Second, the method itself can be objectively evaluated in terms of its
model's ability to predict held-out data.  Such meta-evaluation can be
performed without expensive and unreliable human judgments.

A temporary disadvantage of this approach is that research on
multitext modeling has not begun yet.  The problem of inducing a PGMTG
can be approached from the perspective of bilingual language modeling
\cite{Wu97}, with at least all the methods and challenges of
monolingual language modeling.  Estimation of a monolingual PGMTG
would be hampered by the relatively small size of suitable training
data.  On the other hand, it is easier to estimate a translation model
from a given language to itself than to other languages, if only
because the identity relation provides an excellent word-to-word
translation model as a starting point.

When good multitext models
become available, generalized parsers will become the engine driving
yet another important part of the standard SMT architecture.

\section{Putting it all together}

\begin{figure*}
\centerline{\psfig{figure=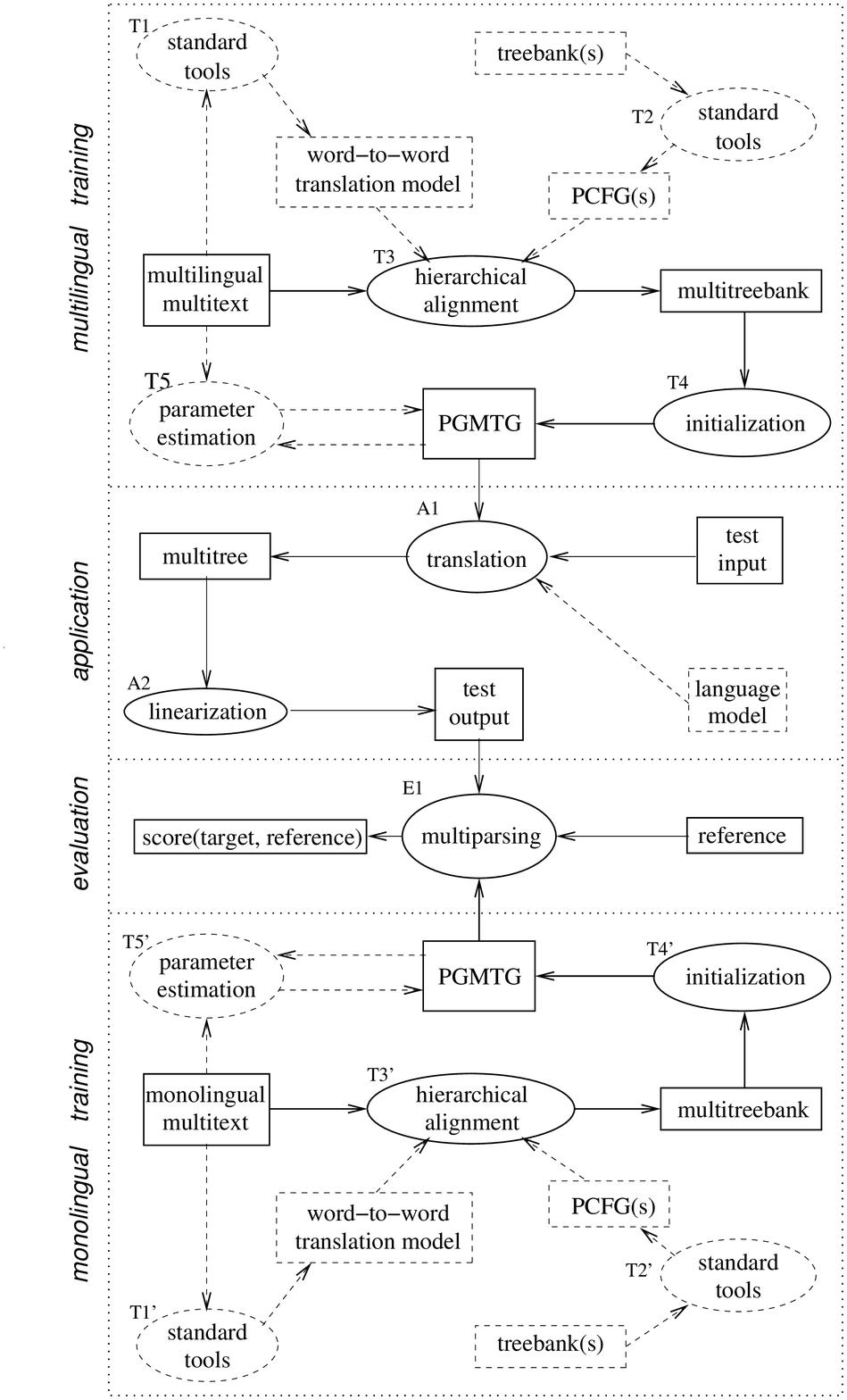,height=8in}}
\caption{Data-flow diagram for a rudimentary system for SMT by
parsing.  Boxes are data; ovals are processes; arcs are flows; dashed
flows and data are recommended but optional.
\label{dfd}}
\end{figure*}

Figure~\ref{dfd} shows the data-flow diagram for a rudimentary SMT system
that is driven by tree-structured translation models.  All the
generalized parsing algorithms involved can be implemented as
different parameterizations of the abstract parsing algorithm in
Section~\ref{sec:apa}.  Below is a sample recipe for running a system
of this kind through training, application to new inputs, and
evaluation.  Unless stated otherwise, each generalized parser's goal
is an item that spans the input and is labeled with the start symbol
of the grammar.  At runtime, the abstract parser's termination
conditions would typically involve goal items as well as limits on
time and/or memory consumption. Any search strategy can be used, at
least in theory. In practice, we must manage computational complexity,
so best-first search is a common favorite.
\be
\item[T1.] Induce a word-to-word translation model.  Use Logic~A
\cite{Melamed03} with enough goal items spanning one word from each
component to cover the input.  Alternatively, publicly available
WFST-based tools can be used \cite{OchNey03}.
\item[T2.] Induce PCFG(s) from monolingual treebank(s), e.g.\
by computing the relative frequencies of productions.
\item[T3.] Hierarchically align the training multitext, e.g.\ using
Logic~C and the derivation-forest semiring.  Constraining PCFG(s) and a
word-to-word translation model can be used to imitate a PGMTG, as
described in Section~\ref{sec:align}.  Other approximations can be
used if these knowledge sources are not available or if other relevant
knowledge sources are available.
\item[T4.] Induce an initial PGMTG from the multitreebank, e.g.\ by computing
 the relative frequencies of productions.
\item[T5.] Re-estimate the PGMTG parameters using Logic~CR, starting
with the initial PGMTG.  Ideally, use the inside semiring, but if
that's too expensive, then use Viterbi-$n$-best.  In addition to the
usual goal item, the termination condition involves the reverse item
corresponding to each of the input axioms.
\item[T1'-T5'] Same as T1-T5, but starting with monolingual
multitext.  The identity relation can be used for T1' as a short-cut.
\item[A1.] Use the PGMTG to infer the most probable multitree covering
the input multitext.  Use Logic~CT under the Viterbi-derivation
semiring.  If a target language model is available, use Logic~CTM
\cite{Melamed04b}.
\item[A2.] Linearize the output yield of the multitree.
\item[E1.] For each component of the test output, multiparse the
bitext consisting of this component and the corresponding reference
translation, using Logic~C under the inside semiring and the
monolingual PGMTG.
\ee
A variety of algorithms have been proposed for Process~T1
\cite{Melamed00,OchNey03} and some of them are available as free
software. Processes~T2, T4, T2', T4', and~A2 are
trivial. Processes~T3, T5, T3', T5', A1, and~E1 are the
generalizations of parsing and their applications presented in this
article.  The ``Statistical Machine Translation by Parsing'' team at
the 2005 JHU Language Engineering Workshop used this recipe to build
GenPar, the first publicly available system of this type
\cite{Burbank+05}.  GenPar revolves around a single abstract parser.

\section{Summary and Outlook}

This article has extended the theory of semiring parsing to present a
new analysis of many common parsing algorithms, as well as other
algorithms that are not usually considered parsing algorithms.  The
analysis revealed that all of these algorithms can be implemented by
an abstract parsing algorithm with five functional parameters: a
grammar, a logic, a semiring, a search strategy, and a termination
condition.  The article then varied two of these functional parameters
--- the logic and the grammar --- to arrive at the class of
translators and the class of hierarchical aligners.  In this manner,
the article has elucidated the relationships between ordinary parsing
and these other classes of algorithms more precisely than previously
possible.  The article then presented two new applications of
generalized parsing, and showed how the various generalizations and
their applications can be used to do all the heavy lifting in a
rudimentary system for statistical machine translation by generalized
parsing.

There are distinct advantages to building SMT systems in this manner.
The software engineering advantage is that improvements invented for
one of these algorithms can often be applied to all of them.  For
example, \namecite{Melamed03} showed how to reduce the computational
complexity of a multiparser by a factor of $n^D$, just by changing the
logic.  The same optimization can be applied to any generalized parser
based on Logics~C, CT, or~CR.  
With good software design, such optimizations need never be
implemented more than once.  Researchers who adopt this approach
can concentrate their talents on better models, without worrying about
system-specific ``decoders.''

A more important advantage in the long term is that this approach to
building MT systems encourages MT research to be less specialized and
more transparently related to the rest of computational linguistics.
A well-understood connection between parsing and SMT algorithms can
foster a stronger connection between research in SMT and research in
the rest of computational linguistics, a connection that has been
weakening in recent years to the detriment of both research
communities.  Research on SMT by Parsing can build on past and future
research on ordinary parsing.  Stronger connections between the two
research communities would enable more researchers to contribute to MT
research, accelerating progress. Conversely, we expect generalized
parsers to be useful for other problems with a similar structure, such
as sentence compression \cite{KnightMarcu00} and structured generation
\cite{Langkilde00}.

The viability of statistical machine translation by generalized
parsing will hinge on development of more powerful logics and grammar
formalisms than the simplistic examples used in this article.
Improved machine learning methods will also be critical.  We
conjecture that the best SMT systems of the near future will combine
new learning algorithms with the expressive power of tree-structured
translation models.  However, inference is likely to remain the main
source of complexity, both conceptual and computational, in these new
learning algorithms.  As better parameters are invented for the
abstract parsing algorithm, we expect the abstractions presented in
this article to become even more important in reducing the complexity
of statistical machine translation systems.

\starttwocolumn

\section*{Acknowledgments}
This article greatly benefited from discussions with Giorgio Satta,
Joseph Turian, and the ``Statistical Machine Translation by Parsing''
Workshop Team: Andrea Burbank, Marine Carpuat, Stephen Clark, Markus
Dreyer, Pamela Fox, Declan Groves, Keith Hall, Mary Hearne, Yihai
Shen, Andy Way, Ben Wellington, and Dekai Wu.  Thanks also to the
anonymous reviewers for helpful feedback.  This research was supported
by an NSF CAREER Award, by the DARPA TIDES program, and by an
equipment gift from Sun Microsystems.


\bibliographystyle{fullname}
\bibliography{article}

\end{spacing}

\end{document}